\documentclass[sigplan,screen]{acmart}

\startPage{1}

\usepackage{booktabs}   
\usepackage{subcaption} 
\usepackage{balance}
\usepackage{booktabs}
\usepackage{pifont}
\usepackage{rotating}
\usepackage{amsfonts}
\usepackage{multirow}
\usepackage{color}
\usepackage{colortbl}
\usepackage{graphicx}

\usepackage{amsmath}
\usepackage{algpseudocode}
\usepackage{algorithmicx}
\usepackage{algorithm} 
\usepackage{subcaption}
\usepackage{wrapfig}
\usepackage{float}

\usepackage{enumitem}
\usepackage{xspace}
\usepackage[normalem]{ulem}
\usepackage{listings}
\usepackage{pgfplots}
\usepackage{pgfplotstable}
\usetikzlibrary{arrows.meta}
\usepackage{relsize}

\tikzset{fontscale/.style = {font=\relsize{#1}}
}
\usepackage{tikz}
\usetikzlibrary{positioning, shapes}
\usetikzlibrary{automata,positioning}

\newcommand{\DefMacro}[2]{\expandafter\newcommand\csname rmk-#1\endcsname{#2}}
\newcommand{\UseMacro}[1]{\csname rmk-#1\endcsname}

\newcommand{\CodeIn}[1]{\begin{small}\texttt{#1}\end{small}}
\newcommand{\XComment}[1]{}
\newcommand{\Space}[1]{}

\newcommand{\Caption}[1]{\caption{#1}\vspace*{-2ex}}

\definecolor{gray}{RGB}{211,211,211}
\newcommand{\jbasicstyle}{\small\sffamily}

\newcommand{\jnumberstyle}{\scriptsize}

\usepackage[textwidth=60pt,textsize=scriptsize]{todonotes} 

\lstdefinelanguage{pseudo}
{ morekeywords={for, in, break, continue, try, except, not,
  if,else,return,map,fieldElement_array_array40,fieldElement_array40},
  keywordstyle=\bfseries, lineskip=-0.1em, numbers=left,
  numberstyle=\jnumberstyle, numbersep=4pt, basicstyle=\jbasicstyle,
  breaklines=true, breakautoindent=true, tabsize=2,
  columns=fullflexible, morecomment=*[l][\textsl]{//},
  mathescape=true, }
\definecolor{javared}{rgb}{0.6,0,0} 
\definecolor{javagreen}{rgb}{0.25,0.5,0.35} 
\definecolor{javapurple}{rgb}{0.5,0,0.35} 
\definecolor{javadocblue}{rgb}{0.25,0.35,0.75} 
 
\newcommand{\NumProperties}{16}
\newcommand{\NumModels}{6}

\newcommand{\AccMC}{\emph{Acc$_{MC}$}}
\newcommand{\DiffMC}{\emph{Diff$_{MC}$}}

\newcommand{\RQOneLower}{How effective are ML models in learning relational properties?}
\newcommand{\RQTwoLower}{How well do ML (decision tree) models generalize (outside of the test set)?}
\newcommand{\RQThreeLower}{How do symmetries in the dataset affect performance of the ML models?}
\newcommand{\RQFourLower}{How does mismatch in symmetries present in the training and evaluation sets affect the performance?}
\newcommand{\RQFiveLower}{What is the quantitative difference between two decision trees models?}

\newcommand{\RQOne}{RQ1: How Effective are ML Models in Learning Relational Properties?}
\newcommand{\RQTwo}{RQ2: How Well do ML (Decision Tree) Models Generalize (Outside of the Test Set)?}
\newcommand{\RQThree}{RQ3: How do Symmetries in the Dataset Affect Performance of the ML Models?}
\newcommand{\RQFour}{RQ4: How does Mismatch in Symmetries Present in the Training and Evaluation Sets Affect the Performance?}
\newcommand{\RQFive}{RQ5: What is the Quantitative Difference Between Two Decision Tree Models?}
\newcolumntype{L}[1]{>{\raggedright\let\newline\\\arraybackslash\hspace{0pt}}m{#1}}
\newcolumntype{C}[1]{>{\centering\let\newline\\\arraybackslash\hspace{0pt}}m{#1}}
\newcolumntype{R}[1]{>{\raggedleft\let\newline\\\arraybackslash\hspace{0pt}}m{#1}}

\newenvironment{CodeOut}{\begin{small}}{\end{small}}

\copyrightyear{2020}
\acmYear{2020}
\setcopyright{acmlicensed}\acmConference[PLDI '20]{Proceedings of the 41st ACM SIGPLAN International Conference on Programming Language Design and Implementation}{June 15--20, 2020}{London, UK}
\acmBooktitle{Proceedings of the 41st ACM SIGPLAN International Conference on Programming Language Design and Implementation (PLDI '20), June 15--20, 2020, London, UK}
\acmPrice{15.00}
\acmDOI{10.1145/3385412.3386015}
\acmISBN{978-1-4503-7613-6/20/06}

\begin{document}

	\title{A Study of the Learnability of Relational Properties}         
	\subtitle{Model Counting Meets Machine Learning (MCML)}                     

	
\author[Usman]{Muhammad Usman}
\affiliation{
	\institution{University of Texas at Austin, USA}            
}
\email{muhammadusman@utexas.edu}          

\author[Wang]{Wenxi Wang}
\affiliation{
	\institution{University of Texas at Austin, USA}            
}
\email{wenxiw@utexas.edu}          

\author[Vasic]{Marko Vasic}
\affiliation{
	\institution{University of Texas at Austin, USA}            
}
\email{vasic@utexas.edu}          

\author[Wang]{Kaiyuan Wang}
\authornote{Research performed while at the University of Texas at Austin.}
\affiliation{
	\institution{Google Inc., USA}            
}
\email{kaiyuanw@google.com}          

\author[Vikalo]{Haris Vikalo}
\affiliation{
	\institution{University of Texas at Austin, USA}            
}
\email{hvikalo@ece.utexas.edu}          

\author[Khurshid]{Sarfraz Khurshid}
\affiliation{
	\institution{University of Texas at Austin, USA}            
}
\email{khurshid@ece.utexas.edu}          

	\begin{abstract}
		This paper
introduces the \emph{MCML} approach for empirically studying
the \emph{learnability} of \emph{relational} properties that can be
expressed in the well-known software design language \emph{Alloy}.  A
key novelty of MCML is quantification of the performance of and semantic
differences among trained machine learning (ML) models, specifically
decision trees, with respect to \emph{entire} (bounded) input spaces,
and not just for given training and test
datasets (as is the common practice).  MCML reduces the quantification
problems to the classic complexity theory problem of \emph{model
counting}, and employs state-of-the-art model
counters.  The results show that relatively simple
ML models can achieve surprisingly high performance (accuracy and F1-score)
when evaluated in the common setting of using
training and test datasets -- even when the training dataset is
much smaller than the test dataset -- indicating the seeming
simplicity of learning relational properties.  However, MCML
metrics based on model counting show that the performance can degrade
substantially when tested against the entire (bounded) input space,
indicating the high complexity of precisely learning these properties,
and the usefulness of model counting in quantifying the true performance.

 
	\end{abstract}



	\maketitle
	\section{Introduction}
	
\emph{Relational properties} which relate abstract entities that can be
viewed as vertices in a graph via edges that define the relations,
e.g., the connectivity of a social network, or
of an object graph on the heap, offer various benefits in the
development of software
systems~\cite{Jackson01AlloyAlpha,RumbaughETAL98UML,Spivey92Z}.  For
example, they enable automated analyses of software requirements,
designs, specifications, and implementations~\cite{10.1007/978-3-642-24372-1_5}.  However, in
most cases, to benefit from the analyses the properties must be
written manually.  For complex systems~\cite{ZaveChrod2012,Sangal:2005:UDM:1103845.1094824},
writing them correctly is challenging, and faults in the properties'
statements can lead to erroneous confidence in correctness.

Our motivation is to leverage advances in machine learning (ML) to
enhance software analyses, thereby improving software quality and
reducing the cost of software failures.  Our focus is on training
binary classifiers that characterize relational properties.  Such
classifiers, once trained to have high accuracy, can
offer much value in automated software analysis.  For example, an
executable classifier can serve as a \emph{run-time check}, e.g., in
an assertion, to validate that the program states at that point
conform to the property represented by the classifier.  Moreover,
executable checks enable automated test input
generation~\cite{57624,Boyapati:2002:KAT:566172.566191}, static
analysis~\cite{Garg:2016:LIU:2837614.2837664,Si:2018:LLI:3327757.3327873},
error
recovery~\cite{DBLP:conf/oopsla/DemskyR03,Ke:2015:RPS:2916135.2916260},
and automated theorem proving~\cite{MouraKADR15}.

ML models can also be utilized in tandem with program synthesis
techniques such as those based on \emph{sketching}~\cite{Sketching},
e.g., a decision tree can provide the basis for a sketch that is
completed by synthesis, or \emph{holes} in a sketch can be filled in
by the decision tree logic~\cite{GopinathETALICSE2014}.  Moreover, learnt properties, even when
somewhat imprecise, can be useful for the developer, e.g., a decision
tree that approximates the properties of program states at a specific
control point can provide insight into likely program behaviors and
help seed other analyses.

In this paper, we introduce the \emph{MCML} approach for empirically
studying the \emph{learnability} of a key class of relational
properties that can be written in the well-known software design
language \emph{Alloy}~\cite{Jackson01AlloyAlpha}.  Our aim is not to
formalize learnability~\cite{Blumer:1989:LVD:76359.76371} of
relational properties; rather, we aim to perform controlled
experiments and rigorously study various well-known properties over
small relations and graphs in order to gain insights into the
potential role of ML methods in this important domain.  Specifically,
we consider training binary classifiers with respect to relational
properties such that the trained classifiers accurately represent the
properties, e.g., training a decision tree classifier to
accurately classify each
input as a directed-acyclic graph (DAG) or not a DAG.

A key novelty of MCML is that it allows quantifying the performance of
trained decision tree models with respect to the \emph{entire} input
space (for a bounded universe of discourse) by utilizing given ground
truth formulas, thereby enabling an evaluation of learnability that is
not feasible using the commonly used ML approaches based on training
and test datasets alone.  Likewise, MCML allows quantifying
differences among trained decision tree models for the entire
(bounded) input space.

To quantify model performance and differences --
irrespective of the datasets -- MCML reduces the quantification
problems to the classic complexity theory problem of \emph{model
counting}, which is to compute the number of solutions to a logical
formula~\cite{Gomes08modelcounting}.  The formulas in our case
represent semantic differences between the ground truth and the
trained model, or between the two models.  Given ground truth $\phi$
and decision tree $d$, the \emph{false negative} count for $d$ is the
model count of ``$\phi\wedge\psi$'' where $\psi$ is $d$'s logic that
leads to the output ``0'' because any solution to ``$\phi\wedge\psi$''
conforms to the ground truth but leads the decision tree to output
``0''; likewise, the \emph{false positive} count for $d$ is the model
count of ``$\neg\phi\wedge\tau$'' where $\tau$ is $d$'s logic that
leads to the output ``1''.  The \emph{true positive} and \emph{true
negative} counts are defined similarly. Using these counts accuracy,
precision, recall and F1-score can be derived.  Furthermore, given two
decision trees $d_1$ and $d_2$, their \emph{semantic difference},
i.e., the number of inputs for which the tree outputs differ, is the
sum of the model count of ``$\psi_1\wedge\tau_2$'' and the model count
of ``$\tau_1\wedge\psi_2$'', where $\psi_i$ is $d_i$'s logic that
outputs ``0'', and $\tau_i$ is $d_i$'s logic that outputs ``1''
($i\in\{1, 2\}$).

Model counting generalizes the propositional satisfiability checking
(SAT) problem, and
is \#P-complete~\cite{Sahni:1976:PAP:321958.321975}.  Despite the
theoretical complexity, recent technological and algorithmic advances
have led to practical tools that can handle very large state
spaces~\cite{MiniSAT2004,ProcedingsSAT2018}.  To embody MCML and quantify the performance
and differences by means of model counting we employ two
state-of-the-art tools: ApproxMC, which uses approximation
techniques to estimate the number of solutions with high precision and
provides formal guarantees~\cite{SM19}; and
ProjMC~\cite{ProjMC}, which uses an effective disjunctive decomposition
scheme to compute the exact number of solutions.  As is common with
many tools for propositional logic, ApproxMC and ProjMC take as
input propositional formulas in conjunctive normal form (CNF) -- the
standard input format for SAT solvers.  To create model counting
problems in CNF, we define a translation from decision trees to CNF in
the spirit of previous work~\cite{article1987}.  The translation
creates succinct CNF formulas that are linear in the size of the input
trees.

To train ML models, we use the standard practice of employing training
datasets.  To evaluate the trained models, we use both the standard
practice of employing test datasets and the MCML metrics.  Our study
has three key distinguishing characteristics: 1)~for each property, we
use \emph{bounded exhaustive} sets of positive samples which
contain \emph{every} positive sample (up to partial symmetry breaking)
within a bounded universe of discourse; 2)~we leverage ground truth
formulas to quantify the performance of trained decision trees with
respect to the \emph{entire} input space, not only the given
datasets; and 3)~we quantify semantic differences among
different trained decision trees with respect to the entire
input space without the use of ground truth formulas.  Moreover, we
evaluate different strategies for splitting the datasets into training
sets and test sets, including ratios where the amount of training data
is much smaller than the amount of test data.

To create the datasets for learning, we rely on logical formulas that
describe the relational properties in Alloy -- a
first-order language with transitive
closure~\cite{Jackson01AlloyAlpha}.  For each property, we use
Alloy's SAT-based back-end to enumerate all the
solutions, i.e., valuations that exhibit the property, up to Alloy's
default \emph{symmetry breaking} which heuristically removes many but
not all isomorphic solutions.  The solutions created by the Alloy
analyzer serve as the samples for training and
test/evaluation.  The solution spaces are very large
-- even with small bounds on the number of entities in the
relations.  For non-trivial properties, the number of positive samples
is far smaller than the number of negative
samples, and exhaustive enumeration of all negative samples is
intractable.  To avoid erroneously biasing the ML models to simply
predict false if the datasets overwhelmingly consist of negative
samples, we create \emph{balanced} sets that contain the same number
of positive and negative
samples~\cite{doi:10.1162/evco.2009.17.3.275}.

As subjects, we use six~ML models, including decision trees,
SVMs, and multi-layer perceptrons, and train them
using datasets from \NumProperties~relational properties over small
relations and graphs.  We use the adjacency matrix representation for
each data item in the training and test datasets; for
example, for a relation over 7~entities, i.e., a graph with
7~vertices, we use 49 boolean inputs for the binary classifier which
outputs true or false as the predicted value, and the space of all
possible inputs has $2^{49}$ elements.

The results show that relatively simple ML models can achieve
surprisingly high performance (accuracy and F1-score) at learning relational properties when
evaluated in the common setting of using training and test datasets --
even when the training dataset is substantially smaller than the test
dataset -- indicating the seeming simplicity of learning these
properties.  However, the use of MCML metrics based on model counting
shows that the performance can degrade substantially when tested against
the entire (bounded) input space, indicating the high complexity of
precisely learning these properties, and the usefulness of model
counting in quantifying the true performance.

The contributions of this paper are as follows.
1)~\emph{Learning relational properties}.  We present a systematic
study of learning \NumProperties~relational properties using
\NumModels~off-the-shelf machine learning models;
2)~\emph{Model counting to quantify performance}.  We reduce the problem
  of evaluating the performance of trained decision tree models over
  the entire input space with respect to ground truth formulas to the
  problem of model counting, and employ cutting edge approximate and
  exact counters to embody the reduction; and
3)~\emph{Model counting to quantify semantic differences}.  We also
  introduce the use of model counting to quantify semantic differences
  between trained decision trees over the entire input space --
  without the need for ground truth formulas or evaluation datasets.

We believe the use of model counting in learning is a promising
research area that can help gain deeper insights into the trained
models, which can inform practical decisions on how to best utilize
the models.  For example, if a trained model in a deployed system is
to be upgraded to a more sophisticated model, model counting could be
a metric that in part informs the decision to upgrade.



	\section{Related Work}
\label{sec:related}
To our knowledge, MCML is the first work to introduce the use of model
counting to quantify performance of trained decision trees with
respect to ground truth formulas and to quantify semantic differences
among different decision trees.

\noindent{\bf Model Counting Applications in ML}.
The use of model counting in machine learning has focused largely on
probabilistic
reasoning~\cite{Chavira2008OnPI,FierensETAL2012,GensDomingos2013,LiangETAL2017UAI}.
Recent work by Baluta ~\cite{Baluta2019quantitative} introduced
model counting for quantifying differences between binarized neural
networks~\cite{hubara2016binarized} that admit a translation to
SAT/CNF~\cite{NarodytskaKRSW18}.  This translation enables our MCML
metrics to generalize beyond decision trees and become applicable to
quantify the performance of binarized neural networks with respect to
the entire input space.  The MCML metrics also directly generalize to
other techniques that use different solvers for model counting, e.g.,
techniques based on ODDs and
OBDDs~\cite{ChanDarwiche2003,ShihETAL2018}.

\noindent{\bf Verification and Testing of ML models}.
Verification and
testing of machine learning models is an active area of research,
including work on novel decision procedures such as
Reluplex~\cite{HuangKWW17,KaBaDiJuKo17Reluplex}, which has been
optimized for the analysis of neural networks with ReLU activation
functions, testing trained models~\cite{PeiETAL17,TianETAL18DeepTest},
applying symbolic
execution~\cite{SunETAL2018ASE,GopinathETAL2018Arxiv}, and
inferring verifiable policies by mimicking deep reinforcement learning
agents~\cite{BastaniETAL18VerifiableRL,VasicETAL19MOET}.  A key difference between
MCML and previous work on verification of properties of trained models
is the focus of previous work on either verifying properties of one
trained model, or checking equivalence or implication between two
models.  In contrast, MCML metrics apply in a more general setting, even when two models are neither equivalent nor such that one implies the other.  

For translating decision trees to CNF formulas,
Hastad~\cite{article1987} introduced the idea to represent a decision
tree as a Disjunctive Normal Form (DNF) formula.  He also
showed that these DNF formulas are convertible to Conjunctive Normal
Form (CNF) formulas.  MCML leverages this work and uses CNF formulas
of trained decision trees to create the quantification problems.
While the formulas MCML creates are optimal in terms of the size of
the CNF formula with respect to the given input tree in the general
case, an alternative approach is to use re-writing, e.g., aka
compilation~\cite{ChanDarwiche2003,ShihETAL2018}, to create smaller
CNF formulas in some specific cases.  However, re-writing itself has a
cost, which can be substantial for non-trivial formulas, e.g., with
hundreds of variables (as for our subjects).  Moreover, when
compilation transforms a decision tree to a simpler decision tree,
MCML works \emph{in tandem} with compilation: first apply re-writing
to reduce the tree, and then use MCML's translation to create a
reduced CNF formula.

\noindent{\bf Analyzing Learnability}.
Efforts to understand the ability of a machine learning model to
generalize are at the core of statistical learning theory.  The key
concepts for establishing such results include the Probably
Approximately Correct (PAC) learning framework
\cite{Valiant:1984:TL:1968.1972}, Vapnik- Chervonenkis (VC) theory
\cite{article1971a}, and the general learning setting
\cite{Vapnik:1995:NSL:211359,Shalev-Shwartz:2010:LSU:1756006.1953019}.
These techniques enable formal analytical characterization of the
number of examples needed to train models for binary classification
tasks and may provide useful intuition and offer guidance about the
design of learning algorithms.
Blumer~\cite{Blumer:1989:LVD:76359.76371} showed that the finiteness
of Vapnik-Chervnenkis (VC) dimension is a basic requirement for
distribution-free learning.  In particular, by relying on the VC
dimension, he analyzed the performance in terms of closure and
complexity and provided a detailed set of conditions for
learnability.

The PAC learnability concepts, however, provide limited insight in the
performance of methods for learning relational properties.  This is
because the number of positively labeled samples in the domain set
(i.e., the space of relational properties) is often orders of
magnitude smaller than the number of negatively labeled ones.
Therefore, Precision, Recall and F1-score, formally defined in
Section~\ref{sec:study}, are rather more informative performance
metrics than the average 0-1 loss; the latter is the focus of the PAC
learnability analysis. In contrast, MCML can precisely quantify
generalizability both in terms of the accuracy (i.e., the 0-1 loss) as
well as precision, recall and F1-score with respect to a given ground
truth.

Indeed, our study complements existing theoretical frameworks by
providing insights based on empirical evidence from controlled
experiments using various relational properties that are common in
software systems.

\noindent{\bf Learning Program Properties}.
In the context of learning properties of code, specifically of dynamic
data structures in Java programs, two recent
projects~\cite{MolinaETAL2019ICSE,UsmanETAL2019SPIN} used a variety of
machine learning models to show the effectiveness of off-the-shelf
models.  However, both these projects used the traditional metrics
with training and test datasets, and did not evaluate the performance
of the trained models with respect to the entire (bounded) input
spaces.  More broadly, machine learning enabled automated
detection of program errors and repair of
faults~\cite{1317470,LongRinardPOPL2016}.

\noindent{\bf Alloy}.
Alloy has been used in several projects: for design and modeling of
software
systems~\cite{JacksonS00,KhurshidJackson00ExploringDesignIntentional,ZaveChordTSE,BagheriETAL2018,WickersonETAL2017,ChongETAL2018};
for software analyses, including deep static
checking~\cite{JacksonVaziri00Bugs,TACOGaleottiETALTSE2013},
systematic testing~\cite{MarinovKhurshid01TestEra}, data structure
repair~\cite{SamimiETALECOOP2010,ZaeemKhurshidECOOP2010}, and
automated debugging~\cite{GopinathETALTACAS2011}; for analysis of
hardware
systems~\cite{CheckMateMICRO2018,CheckMateMicro2019,CheckMateGitHub};
and testing and studying model
counters~\cite{TestMC2019,StudySymmetry}.  MCML introduces the use of
Alloy for creating training and test data for machine learning models
and leverages Alloy's backend for creating CNF formulas that represent
the ground truth.


	\begin{figure}[!t]
\begin{CodeOut}
\begin{verbatim}
sig S { r: set S } // r is a binary relation of type SxS
pred Reflexive() { all s: S | s->s in r }
pred Symmetric() {
  all s, t: S | s->t in r implies t->s in r }
pred Transitive() { all s, t, u: S |
  s->t in r and t->u in r implies s->u in r }
pred Equivalence() {
  Reflexive and  Symmetric and Transitive }
E4: run Equivalence for exactly 4 S
\end{verbatim}
\end{CodeOut}
\vspace*{-2ex}
\caption{Alloy specification with one set $S$, one binary relation
  $r$, four predicates ($Reflexive$, $Symmetric$, $Transitive$, and
  $Equivalence$), and one command ($E4$).\label{fig:example}}
\vspace*{-2ex}
\end{figure}

\section{Example}
\label{sec:example}

Figure~\ref{fig:example} shows an Alloy specification that declares a
set (\CodeIn{sig}) $S$ of atoms, a binary relation $r: S\times S$, and
four predicates (i.e., formulas) that specify reflexive,
symmetric, transitive, and equivalence relations.  The keyword
``\CodeIn{all}'' is universal quantification, ``\CodeIn{in}''
is subset, ``\CodeIn{and}'' is conjunction, and
``\CodeIn{implies}'' is implication.  The operator
`\CodeIn{->}' is Cartesian product; for scalars $s$ and $t$,
$s$\CodeIn{->}$t$ is the pair ($s, t$).  The command
``\CodeIn{E4: run Equivalence for exactly 4 S}'' instructs the Alloy analyzer to solve
the formula(s) that define
equivalence relation with respect to a \emph{scope}, i.e., bound, of
exactly 4~atoms in set \CodeIn{S}.   The analyzer uses the bound to translate the Alloy
specification to a propositional satisfiability formula, and uses
off-the-shelf SAT solvers to solve it.  The Alloy analyzer
supports \emph{incremental} solvers that can
enumerate all solutions.

\begin{figure}[!t]
	\includegraphics[width=1.8in]{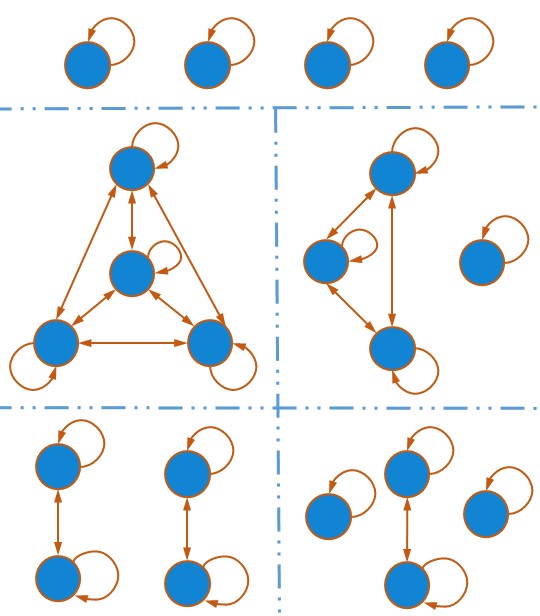}
	\vspace*{-2ex}
	\caption{5 non-isomorphic equivalence
                  relations.\label{fig:instances}}
	\vspace*{-4ex}
\end{figure}

Executing the command \CodeIn{E4} and enumerating all solutions
creates the 5~solutions illustrated in Figure~\ref{fig:instances}.
Note, each solution is non-isomorphic.  The Alloy analyzer adds
symmetry breaking predicates during the translation, which breaks
several (but, in general, not all)
symmetries~\cite{Shlyakhter01EffectiveSymmetryBreaking}.  The space of
all candidate solutions for the command \CodeIn{E4} has size
$2^{16}$=65,536~since each candidate is a $4\times 4$~matrix of
boolean variables.  Note, how quickly the state space grows as the number of
vertices increases, e.g., for just 7 vertices, the state space has
size $2^{(7^2)} = 2^{49}$ which is greater than $10^{14}$.

\noindent{\bf ApproxMC}. To illustrate the use of ApproxMC, consider estimating the number of
solutions for the \CodeIn{Equivalence} predicate with scope~20.  The
Alloy command ``\CodeIn{E20: run Equivalence for exactly 20 S}''
defines the constraint solving problem.  The Alloy analyzer translates
this problem to a CNF formula that has 18,666 variables (of which
400~are primary variables) and 27,202 clauses.  ApproxMC solves this
CNF formula in 17.8~seconds and reports an approximate model count of
11,264.  The exact model count, which we calculate using the Alloy
analyzer (which uses an enumerating SAT solver), is 10,946, i.e., the
ApproxMC count is within 3\% error rate.

\noindent{\bf ProjMC}. To illustrate the use of ProjMC, consider
computing the exact model count for the \CodeIn{Equivalence} predicate for
scope~20.  Given the CNF formula for the Alloy command \CodeIn{E20},
ProjMC reports the exact model count of 10,946 in 351.1~seconds.  The
count reported by ProjMC is, as expected, the exact number of solutions
we get using the Alloy analyzer.


	\section{MCML Approach}

Our approach, called \emph{MCML}, introduces the use of model counting
for quantifying performance of decision tree models in machine
learning (ML).  MCML is embodied by two techniques: \AccMC, which
quantifies the performance of trained ML models with respect to ground
truth formulas; and \DiffMC, which quantifies the semantic differences
between two trained models.  Both \AccMC{} and \DiffMC{} compute the
results with respect to the entire input space and do not require any
datasets.

\begin{wrapfigure}{r}{0.2\textwidth}
	\vspace*{-6ex}\hspace*{-6ex}
	\includegraphics[width=1.75in]{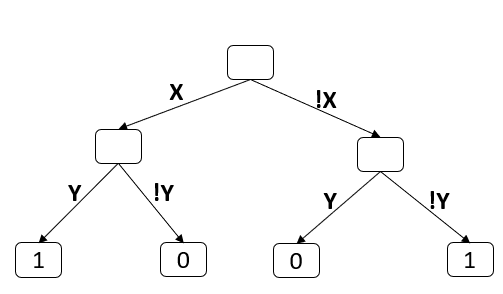}
	\vspace*{-4ex}
	\caption{A decision tree.\label{fig:dtree}}
	\vspace*{-2ex}
\end{wrapfigure}
We view a decision tree as a set of \emph{paths}; any input follows exactly one
path, and each path is a conjunction of \emph{branch conditions} such
that each condition contains one input variable~\cite{doi:10.1080/00224065.1981.11978748}. Since our
focus is on relational properties over graphs that are represented
using adjacency matrices,
the input variables for the decision tree are all binary and so is the
output.
Moreover, each branch condition on any path is simply of the
form either ``$(!v)$'', i.e., input $v$ is ``0'' (\emph{false}), or
``$(v)$'', i.e., input $v$ is ``1'' (\emph{true}).  Therefore, each
branch condition is simply a \emph{literal}, i.e., a variable or its
negation.  Figure~\ref{fig:dtree} illustrates a decision tree with
2~inputs ($x$ and $y$) and 4~paths ([$x,y$], [$x,!y$],
[$!x,y$], and [$!x,!y$]).

\begin{figure}
	\begin{minipage}{\linewidth}
		\centering\captionsetup[subfigure]{justification=centering}
		\includegraphics[width=\textwidth]{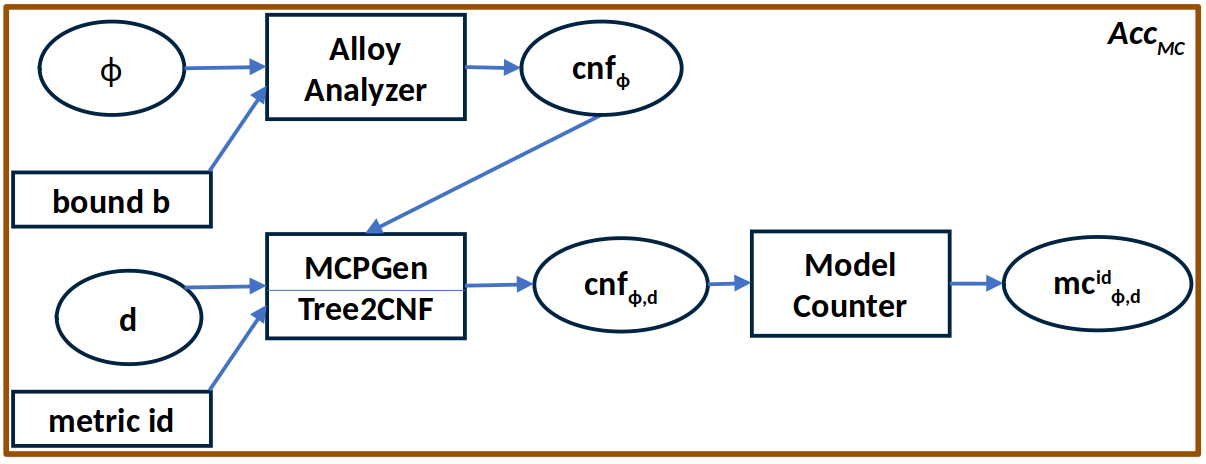}
		\subcaption{Quantifying the performance of decision tree $d$
			w.r.t. ground truth $\phi$.\label{fig:tool-acc}}
		\vspace*{1ex}
		\includegraphics[width=\textwidth]{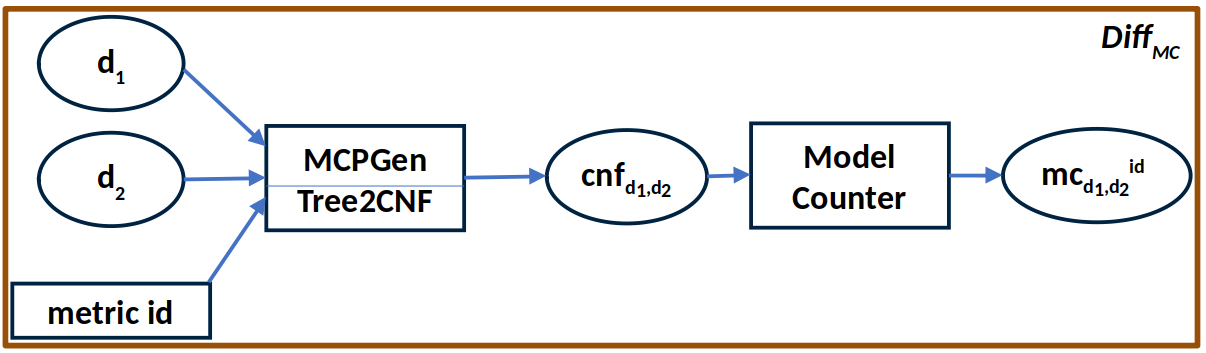}
		\subcaption{Quantifying differences between trees $d_1$ and
			$d_2$.\label{fig:tool-diff}}
	\end{minipage}
        \vspace*{-2ex}
	\caption{MCML Framework.}
        \vspace*{-2ex}
\end{figure}

\label{sec:quant-acc}

\noindent{\bf \AccMC: Quantifying model performance}.
Assume property $\phi$ (ground truth) is learned by a decision tree $d$
that is a binary classifier with $n$ inputs and so the input space has
size $2^n$.  Assume $d$ has $t$ unique paths $p_1, \ldots, p_t$ ($t\ge
0$) that predict label \emph{true}, and $f$ unique paths $q_1, \ldots, q_f$
($f\ge 0$) that predict label \emph{false}.  For path $p$, let $\psi(p)$ be
the conjunction of branch conditions along $p$; we refer to $\psi(p)$ as
the \emph{path condition} for $p$~\cite{pathconditions}.
{}
We define the following four metrics based on model counting to
generalize (for the entire $2^n$ input space) the traditional metrics
of \emph{true positives} ($tp$), \emph{false positives} ($fp$),
\emph{true negatives} ($tn$), and \emph{false negatives} ($fn$), where
$mc(a,b)$ represents the model count for the formula $a$ with respect
to bound $b$:
\\
\vspace*{-5mm}
	\begin{equation}
	tp(\phi,d) = mc(\phi \wedge \bigvee_{i=1}^{t}\psi(p_i), n)\label{eq:one}
\end{equation}
\vspace*{-6.2mm}
\\
	\begin{equation}
	fp(\phi,d) = mc(\neg\phi \wedge \bigvee_{i=1}^{t}\psi(p_i), n)\label{eq:two}
	\end{equation}
	\vspace*{-6.2mm}
\\
	\begin{equation}
	tn(\phi,d) = mc(\neg\phi \wedge \bigvee_{i=1}^{f}\psi(q_i), n)\label{eq:three}
	\end{equation}
	\vspace*{-6.2mm}
\\
	\begin{equation}
	fn(\phi,d) = mc(\phi \wedge \bigvee_{i=1}^{f}\psi(q_i), n)\label{eq:four}
	\end{equation}
\\
\vspace*{-4mm}

The formula $\bigvee_{i=1}^{t}\psi(p_i)$ characterizes the inputs that
the decision tree $t$ classifies as \emph{true}; the formula is a
disjunction of the path conditions for the paths that predict
\emph{true} and any input that is classified as \emph{true} must be a
solution to exactly one of these paths. For example, for the decision tree depicted in Figure~\ref{fig:dtree}, the path condition formula for label \emph{true} should be $(x \wedge y) \vee (!x \wedge !y)$.  Likewise, the formula
$\bigvee_{i=1}^{f}\psi(q_i)$ characterizes the inputs that the
tree $t$ classifies as \emph{false}.

While a direct application of our metrics requires the ground truth
formula $\phi$ that characterizes the property of interest, the
metrics are also applicable when $\phi$ is not known.  Specifically,
the metrics naturally generalize to allow quantifying
\emph{differences} in two trained models, again without requiring any
datasets for quantification.

\label{sec:quant-diff}
\noindent{\bf \DiffMC: Quantifying model differences.}
Let $d_1$ and $d_2$ be decision trees that are trained as binary classifiers (using the same
or different datasets) with $n$ inputs.
Let $d_1$ have paths $p_1, \ldots, p_{t_1}$ ($t_1\ge 0$) that
predict $true$ and paths $q_1, \ldots, q_{f_1}$ ($f_1\ge 0$) that
predict $false$.  Let $d_2$ have paths $r_1, \ldots, r_{t_2}$ ($t_2\ge
0$) that predict $true$ and paths $s_1, \ldots, s_{f_2}$ ($f_2\ge 0$)
that predict $false$.  The following four metrics quantify the number
of inputs (in the entire $2^n$ input space) for which $d_1$ and
$d_2$ make the same decision -- $tt$, i.e., both predict true,
or $ff$, i.e., both predict false -- and numbers of inputs for which
the decisions differ -- $tf$, i.e., $d_1$ predicts $true$ but
$d_2$ predicts false, and $ft$, i.e., $d_1$ predicts $false$ but $d_2$
predicts $true$:
\vspace*{-3mm}
\\
	\begin{equation}
		tt(d_1, d_2) = mc(\bigvee_{i=1}^{t_1}\psi(p_i) \wedge \bigvee_{j=1}^{t_2}\psi(r_j), n)\hspace*{-3ex}
	\end{equation}
\vspace*{-3mm}
\\
	\begin{equation}
tf(d_1, d_2) = mc(\bigvee_{i=1}^{t_1}\psi(p_i) \wedge \bigvee_{j=1}^{f_2}\psi(s_j), n)\hspace*{-3ex}
	\end{equation}
\vspace*{-3mm}
\\
	\begin{equation}
	ff(d_1, d_2) = mc(\bigvee_{i=1}^{f_1}\psi(q_i) \wedge \bigvee_{j=1}^{f_2}\psi(s_j), n)\hspace*{-3ex}
	\end{equation}
	\\
\vspace*{-3mm}
	\begin{equation}
		ft(d_1, d_2) = mc(\bigvee_{i=1}^{f_1}\psi(q_i) \wedge \bigvee_{j=1}^{t_2}\psi(r_j), n)\hspace*{-3ex}
	\end{equation}
\vspace*{-3mm}
\\
\\
The semantic \emph{difference} ($diff$) in $d_1$ and $d_2$ is
quantified as the ratio of the number of inputs $x$ such that
decisions for $x$ differ between $d_1$ and $d_2$, to the total number
of inputs; and the \emph{similarity} ($sim$) in $d_1$ and
$d_2$ is quantified as the ratio of the number of inputs $x$ such that
predicted labels for $x$ match between $d_1$ and $d_2$, to the total
number of inputs:
\\
\vspace*{-2.5mm}
\begin{equation}
diff(d_1, d_2) = \frac{tf(d_1, d_2) + ft(d_1, d_2)}{2^n}\hspace*{-0ex}
\end{equation}
\vspace*{-2.5mm}
\begin{equation}
sim(d_1, d_2) = \frac{tt(d_1, d_2) + ff(d_1, d_2)}{2^n} \hspace*{-1.5ex} 
\end{equation}
\vspace*{-2.5mm}
\begin{equation}
\text{i.e.,}~sim(d_1, d_2) = 1-diff(d_1, d_2) \hspace*{4ex} 
\end{equation}
\vspace*{-2.5mm}
\\

\noindent{\bf Framework}.  We embody the metrics into a prototype framework that translates
decision trees to CNF and leverages off-the-shelf model counters.
Figure~\ref{fig:tool-acc}
illustrates the key steps to quantify the performance of a trained
decision tree model $d$ with respect to the ground truth $\phi$
written in Alloy.  $\phi$ is translated by the Alloy analyzer with
respect to bound $b$ (e.g., 20~vertices) into a CNF formula
$cnf_{\phi}$.  The module \emph{MCPGen} takes as input $d$ and
$cnf_{\phi}$, translates the relevant parts of $d$ with respect to the
desired metric ($tp$, $fp$, $tn$, and $fn$) into CNF using the
sub-module \emph{Tree2CNF}, and outputs the CNF formula $cnf_{\phi,d}$
which defines the model counting problem.  The CNF is an input to
the model counter (e.g., ApproxMC and ProjMC) that outputs the approximate model count and exact model count respectively.
Figure~\ref{fig:tool-diff} illustrates the key steps to quantifying
the semantic differences between two trained decision tree models
$d_1$ and $d_2$.

\noindent{\bf Translating decision tree logic to CNF}.  The goal of
the sub-module \emph{Tree2CNF} is to translate a formula $\rho$ of the form
$\vee_{i=1}^{k}\psi(p_i)$, which represents either all the paths that
predict label \emph{true} or all the paths that predict label
\emph{false}, to CNF.  $\rho$ is originally in \emph{disjunctive
  normal form} and can be translated to CNF using various techniques.
One standard technique is to apply propositional equivalences and De
Morgan's laws~\cite{oldpaper}; however, this technique can lead to a
blow-up in the size of the formula that can negatively impact the
back-end solver's performance.  Another standard technique is to apply
the Tseitin transformation~\cite{Tseytin1966} which creates formulas linear in
the size of the input; however, this technique uses auxiliary
variables and creates an equisatisfiable (but not necessarily
equivalent) formula which can have a different model count than the
original formula.

Our translation uses the following observation~\cite{article1987} that allows
a translation which does not cause a blow up in the formula size and
preserves the model counts: any input that does not get classified as
\emph{true} gets classified as \emph{false} and vice versa.
Therefore, if each $\psi(p_i)$ is a path condition for a path that
leads to label \emph{true} and $\rho=\vee_{i=1}^{k}\psi(p_i)$
represents the decision tree logic that predicts \emph{true}, the
negation of $\rho$ characterizes the decision tree logic that predicts
\emph{false}.  The formula $\neg\rho$ immediately simplifies to CNF
because
$\neg\rho=\neg\vee_{i=1}^{k}\psi(p_i)=\wedge_{i=1}^{k}\neg\psi(p_i)$
and each $\psi(p_i)$ is a conjunction of literals; therefore, each
$\neg\psi(p_i)$ is a disjunction of literals. For example, for the decision tree in Figure~\ref{fig:dtree}, the path condition formula for label \emph{false}  is the negation of the path condition formula for label \emph{true}, that is $!((x \wedge y) \vee (!x \wedge !y))$ which can be directly translated into CNF formula $(!x\vee !y) \wedge (x \vee y)$.

\noindent{\bf Analysis.}  Our translation creates a compact formula
\emph{directly} in CNF without introducing any auxiliary variables --
both for output label~\emph{true} and for label~\emph{false}.  For a decision tree
with $n$ leaf nodes, the number of path conditions (PCs) is $n$, i.e.,
\emph{linear} in the tree size.  If there are $k$ features, the size
of the \emph{CNF formula} is $O(nk)$: each PC has $\le k$ conditions,
each condition is a literal, and the formula is a conjunction of the
negation of each PC.  Thus, in terms of the size of the formula for
the back-end solver, the translation is an optimal choice for CNF with
no auxiliary variables used in the general case.
Our tool embodiment of MCML supports two state-of-the-art back-ends:
1)~the exact model counter ProjMC~\cite{ProjMC}; and 2)~the
approximate model counter ApproxMC~\cite{SM19}.


	\section{Study}
\label{sec:study}

\begin{table*}
\Caption{Subject properties and model counts.  For each property, the
  scope used, the size of the state space, the number of positive
  samples created by Alloy using its default enumeration with symmetry breaking, the count estimated by ApproxMC with symmetry breaking, the
  count estimated by ApproxMC without symmetry breaking, the count computed by projMC with symmetry breaking, and the
  count computed by ProjMC without symmetry breaking are
  shown. ``-'' indicates time-out.\label{tab:datacount}}
\begin{footnotesize}
\begin{tabular}{|@{\hspace{1ex}}l@{\hspace{1ex}}|@{\hspace{1ex}}r@{\hspace{1ex}}|@{\hspace{1ex}}c@{\hspace{1ex}}|@{\hspace{1ex}}r@{\hspace{1ex}}|@{\hspace{1ex}}r@{\hspace{1ex}}|@{\hspace{1ex}}r@{\hspace{1ex}}|@{\hspace{1ex}}r@{\hspace{1ex}}|@{\hspace{1ex}}r@{\hspace{1ex}}|}
\hline
\emph{Property} & \emph{Scope} & \emph{State} & \emph{Valid-} & \emph{Est-Valid-} & \emph{Est-Valid-} & \emph{Valid-} & \emph{Valid-}\\
& & \emph{Space} & \emph{SymBr} &\emph{SymBr} & \emph{NoSymBr} &\emph{SymBr} & \emph{NoSymBr} \\
& & \emph{} & \emph{(Alloy)} &\emph{(ApproxMC)} & \emph{(ApproxMC)} &\emph{(ProjMC)} & \emph{(ProjMC)} \\
\hline
{\emph{Antisymmetric}} & 5 	 &  $2^{25}$		& 56723 & 55296 &  1998848 & 56723 & 1889568
 \\
{\emph{Bijective}}	 & 14    & $2^{196} $ 		& 25043 & 25088 &  - & 25043 &-\\
{\emph{Connex}}& 6     & $2^{36} $				& 148884& 147456 & 14680064  & 148884 & 14348907\\
{\emph{Equivalence}}	 & 20    & $2^{400} $   & 10946 & 11264 &  - & 10946 &-\\
{\emph{Function}}	 	 & 8     & $2^{64} $ 	& 16531 & 16640 &  17563648& 16531 & 16777216
 \\
{\emph{Functional}}	 & 8     &  $2^{64} $ 		& 35017 & 35328 &  48234496& 35017 & 43046721
 \\
{\emph{Injective}}	 & 8 	 &  $2^{64} $ 		& 16531 & 16640 &  17563648& 16531 & 16777216
 \\
{\emph{Irreflexive}}	 & 5     &  $2^{25} $ 	& 35886 & 36352 &  2686976& 35886 &1048576
 \\
{\emph{NonStrictOrder}}& 7     &  $2^{49} $		& 26387 & 26112 &  6422528& 26387 &6129859
 \\
{\emph{PartialOrder}}  & 6     &  $2^{36} $ 	& 82359 & 88064 & 8126464&82359  &8321472
  \\
{\emph{PreOrder}}	 	 & 7     & $2^{49} $ 	& 43651 & 43008 & 9175040  & 43651 &9535241\\
{\emph{Reflexive}}	 & 5 	 & $2^{25} $ 		& 35886 & 35840 & 1048576& 35886 &1048576
  \\
{\emph{StrictOrder}}	 & 7     &  $2^{49} $ 	& 26387 & 29184 & 6815744& 26387 &6129859
  \\
{\emph{Surjective}}	 & 14    &  $2^{196} $ 		& 25043 & 25088 & -  & 25043 &-\\
{\emph{TotalOrder}}	 & 13    &  $2^{169} $ 		& 15511 & 14848 & 5502926848& 15511 &-
  \\
{\emph{Transitive}}	 & 6     & $2^{36} $ 		& 95564 & 102400 & 9306112& 95564 &9415189
\\
\hline
\end{tabular}
\end{footnotesize}
\end{table*}

This section describes our study methodology and summarizes the
evaluation results for the 16~relational properties
(Table~\ref{tab:datacount}) and 6~machine learning models (decision
trees (\emph{DT}), random forest decision trees (\emph{RFT}), Adaboost
decision trees (\emph{ABT}), gradient boosting decision trees
(\emph{GBDT}), support vector machines (\emph{SVM}) and multi-layer
perceptrons (\emph{MLP}) studied.  The properties include:
anti-symmetric, bijective, connex, equivalence, function, functional,
injective, irreflexive, non-strict order, partial order, pre-order,
reflexive, strict order, surjective, total order and transitive.

\noindent{\bf Generation of positive and negative samples}.  For each
property and scope, we use Alloy to create two sets of
solutions -- positive solutions, which satisfy the property, and
negative solutions, which negate it.  For positive solutions, we use
the set of \emph{all} solutions enumerated by Alloy's backend SAT
engine (with respect to the scope).  Due to the complex nature of the
properties, the number of negative solutions is much larger than the
number of positive solutions, and it is infeasible to exhaustively
enumerate all negative solutions.  To create the set of negative
solutions, we sample them at random from the entire state space (with
respect to the scope).  Specifically, to sample a negative solution,
we first create a candidate solution at random and then confirm that
it does not satisfy the property using the Alloy Evaluator, which
simply evaluates the corresponding Alloy formula by replacing all the
variables in the formula with the values in the given candidate
solution, and using constant propagation (without any constraint
solving).

Each data sample has a feature vector and a binary label that is 1 for
the positive class and 0 for the negative class.  The features
linearly represent the adjacency matrix, e.g., if the scope of a
property is 7, a 49-bit feature vector is used.

\noindent{\bf Selection of scope and symmetry breaking}.  We study
each property under two settings for symmetry breaking.  The first
setting is when Alloy's default symmetry breaking constraints are
used; in this setting we choose the smallest scope such that there are
$\ge$10,000 positive solutions.  The second setting is when no
symmetry breaking constraints are added by the Alloy analyzer; in this
setting, we choose the smallest scope such that there are $\ge$90,000
positive solutions.

Table~\ref{tab:datacount}  tabulates for each property, the
scope (i.e., number of atoms in set $S$), the size of the input space, the
number of positive samples created by Alloy, the count of positive
samples estimated by ApproxMC, the count of positive samples
estimated by ApproxMC when symmetry breaking is turned off,  the count of positive samples given by ProjMC and the count of positive samples
given by ProjMC when symmetry breaking is turned off.

\noindent{\bf Training models}.  
A key decision in
  evaluating ML models is to select the training and test ratio. We
use 5~different ratios for \emph{training}:\emph{test}, specifically
75:25, 50:50, 25:75 10:90 and 1:99 to evaluate a wide range of settings.
There is no overlap in the training and test datasets.  We performed
the experiments using basic out-of-the-box models.  We explored tuning
the hyper-parameters but the results were only marginally improved and
the small increase in accuracy was offset by a much larger increase in
time spent on tuning.  We report the results obtained using the basic
models without tuning of hyper-parameters. We used Python programming language and Scikit-Learn Library to implement machine learning models. We kept time-out of 5000 seconds which is common in field of model counting.

\noindent{\bf Performance metrics}.  We used four standard metrics to
determine the quality of classification results: \emph{accuracy},
i.e., $\frac{TP + TN}{TP+FP+TN+FN}$; \emph{precision}, i.e.,
$\frac{TP}{TP+FP}$; \emph{recall}, i.e., $\frac{TP}{TP+FN}$; and
\emph{F1-score}, i.e., $\frac{2*Precision*Recall}{Precision+Recall}$.

Due to space limitations we show results for select properties and
focus on ApproxMC. Complete results of all properties for ApproxMC and ProjMC with respect to all ratios and list of metadata items such as number of
primary variables, total variables and total clauses for each
property are available at : \url{https://github.com/muhammadusman93/MCML-PLDI2020}.  Alloy specifications of the properties and the detailed
results are also available at the GitHub repository.  All
experiments were performed on Ubuntu~16.04 with an Intel Core-i7 8750H
CPU (2.20~GHz) and 16GB~RAM.

\noindent{\bf Research questions.} We answer the following research
questions in our empirical study.
\begin{itemize}
\item{\bf RQ1: }\RQOneLower 
\item{\bf RQ2: }\RQTwoLower 
\item{\bf RQ3: }\RQThreeLower
\item{\bf RQ4: }\RQFourLower 
\item{\bf RQ5: }\RQFiveLower
\end{itemize}

\subsection{Answers to the Research Questions}
\label{sec:eval}
\begin{table}
\Caption{Classification results on the test set for \emph{PartialOrder}. ``\emph{Training:Test}'' ratio is shown in the left column, and default Alloy symmetry breaking is used to create datasets.}
\label{tab:summarypartialon}
\centering
	\begin{footnotesize}
	\begin{tabular}{|@{\hspace{1ex}}r@{\hspace{1ex}}|@{\hspace{1ex}}c@{\hspace{1ex}}|@{\hspace{1ex}}c@{\hspace{1ex}}|@{\hspace{1ex}}c@{\hspace{1ex}}|@{\hspace{1ex}}c@{\hspace{1ex}}|@{\hspace{1ex}}c@{\hspace{1ex}}|}
		\hline
		\emph{Ratio} & \emph{Model}& \emph{Accuracy} & \emph{Precision} & \emph{Recall} & \emph{F1-score}\\
		\hline
{ 75:25 } & DT   &  0.9996  &  0.9992  &  1.0000  &  0.9996  \\
& RFT  &    0.9999  &  0.9998  &  1.0000  &  0.9999  \\
& GBDT  &    0.9951  &  0.9905  &  0.9997  &  0.9951 \\ 
& ABT  &    0.9412  &  0.9396  &  0.9425  &  0.9411\\  
& SVM  &    0.9993  &  0.9985  &  1.0000  &  0.9993  \\
& MLP  &    0.9996  &  0.9993  &  1.0000  &  0.9996  \\ \hline
{ 25:75 } & DT   &  0.9983  &  0.9968  &  0.9999  &  0.9983  \\
& RFT  & 0.9996  &  0.9993  &  1.0000  &  0.9996  \\
& GBDT  &   0.9952  &  0.9906  &  0.9998  &  0.9952 \\ 
& ABT  &   0.9419  &  0.9401  &  0.9441  &  0.9421 \\ 
& SVM  &   0.9987  &  0.9973  &  1.0000  &  0.9987  \\
& MLP  &   0.9991  &  0.9982  &  1.0000  &  0.9991  \\ \hline
{ 1:99 } & DT  &  0.9798  &  0.9656  &  0.9949  &  0.9801\\  
& RFT  &   0.9931  &  0.9937  &  0.9926  &  0.9931  \\
& GBDT  & 0.9904  &  0.9846  &  0.9964  &  0.9904  \\
& ABT  &  0.9387  &  0.9351  &  0.9428  &  0.9389  \\
& SVM  &  0.9769  &  0.9767  &  0.9772  &  0.9770  \\
& MLP  &  0.9909  &  0.9867  &  0.9953  &  0.9910  \\
\hline
\end{tabular}
\end{footnotesize}
\vspace{-8mm}
\end{table}

\begin{table*}
\centering
\Caption{Decision tree performance on the test set (symmetries broken) and in respect to the entire state space encoded by ground truth formula $\phi$ (constrained by symmetry breaking).}
\label{tab:grdon_trainon}
\begin{footnotesize}
\begin{tabular}{|@{\hspace{1ex}}l@{\hspace{1ex}}|@{\hspace{1ex}}r@{\hspace{1ex}}|@{\hspace{1ex}}r@{\hspace{1ex}}|@{\hspace{1ex}}c@{\hspace{1ex}}|@{\hspace{1ex}}c@{\hspace{1ex}}|@{\hspace{1ex}}r@{\hspace{1ex}}|@{\hspace{1ex}}c@{\hspace{1ex}}|@{\hspace{1ex}}c@{\hspace{1ex}}|@{\hspace{1ex}}c@{\hspace{1ex}}|@{\hspace{1ex}}c@{\hspace{1ex}}|}
	\hline
\emph{Property} & \emph{Accuracy} & \emph{Precision} & \emph{Recall} & \emph{F1-score}  & \emph{Accuracy} & \emph{Precision} & \emph{Recall} & \emph{F1-score}& \emph{Time[s]}\\ 
\emph{} & \emph{Test} & \emph{Test} & \emph{Test} & \emph{Test}  & \emph{$\phi$} & \emph{$\phi$} & \emph{$\phi$} & \emph{$\phi$}& \emph{}\\ 
\hline
\emph{Antisymmetric}   &0.9913&0.9871&0.9957&0.9914      	& 	0.9690	 & 0.4576 & 	0.9960 & 	0.6271 &  	0.9 \\ \hline
\emph{Bijective}  	   &0.9971&0.9942&1.0000&0.9971      	 & 0.9729 & 0.0000 & 1.0000 & 0.0000&   219.3\\ \hline
\emph{Connex}          &0.9996&0.9993&1.0000&0.9996      	& 0.9992 & 0.0698 & 1.0000 & 0.1304&   1.3 \\ \hline
\emph{Equivalence}     &0.9958&0.9920&0.9997&0.9958      	 & - & - & 0.9997 & - &   -\\ \hline
\emph{Function}    	   &0.9954&0.9928&0.9980&0.9954      	& 0.9924 & 0.0000 & 0.9982 & 0.0000&  6.0  \\ \hline
\emph{Functional}      &0.9991&0.9983&0.9998&0.9991     	& 0.9943 & 0.0000 & 0.9998 & 0.0000&   6.3 \\ \hline
\emph{Injective}       &0.9979&0.9961&0.9998&0.9980     	& 0.8889 & 0.0000 & 0.9998 & 0.0000&   6.1 \\ \hline
\emph{Irreflexive}     &1.0000&1.0000&1.0000&1.0000         & 1.0000 & 1.0000 & 1.0000 & 1.0000	&  0.4 \\ \hline
\emph{NonStrictOrder}  &0.9994&0.9989&1.0000&0.9994      	& 0.9944 & 0.0000 & 1.0000 & 0.0000&   3.1 \\ \hline
\emph{PartialOrder}    &0.9963&0.9936&0.9990&0.9963      	& 0.9675 & 0.0059 & 0.9991 & 0.0116&  2.3 \\ \hline
\emph{PreOrder}        &0.9992&0.9985&0.9999&0.9992      	& 0.9909 & 0.0000 & 0.9999 & 0.0000&  3.5 \\ \hline
\emph{Reflexive}       &1.0000&1.0000&1.0000&1.0000      	& 1.0000 & 1.0000 & 1.0000 & 1.0000&  0.4 \\ \hline
\emph{Strictorder}     &0.9991&0.9982&1.0000&0.9991      	& 0.9915 & 0.0000 & 1.0000 & 0.0000&  3.0 \\ \hline
\emph{Surjective}      &0.9980&0.9961&1.0000&0.9980      	& 0.9993 & 0.0000 & 1.0000 & 0.0000&  208.1 \\ \hline
\emph{TotalOrder}      &0.9994&0.9988&1.0000&0.9994      	& 	0.9983 & 	0.0000 & 	1.0000 & 	0.0000&  95.7 \\ \hline
\emph{Transitive}      &0.9949&0.9910&0.9989&0.9949      	& 0.9866 & 0.0030 & 0.9990 & 0.0059&   2.1 \\ \hline
\end{tabular}
\end{footnotesize}

\end{table*}

\subsubsection{\RQOne}

Table ~\ref{tab:summarypartialon} summarizes performance of ML classifiers for one selected property (PartialOrder).
The dataset is generated using default Alloy symmetry breaking.
The dataset is further split into training and test datasets using different ratios ($75$:$25$, $25$:$75$ and $1$:$99$) to measure the performance change when using different splits of the datasets.
All models exhibit high accuracy and F1-score,
where the accuracy is in range $[0.94, 1.00]$, the precision is in range $[0.94, 1.00]$, the recall is in range $[0.94, 1.00]$, and the F1-score is in range $[0.94, 1.00]$.
Overall, all models achieve good performance, and surprisingly even for a small training:test ratio ($1:99$) models achieve good performance.
We have also performed similar experiments for all other relational properties (available at GitHub repository),
and results are similar as for the PartialOrder -- accuracy is in range $[0.92, 1.00]$, and the F1-score is in range $[0.92, 1.00]$, where the lowest accuracy is for Antisymmetric property.
It is surprising to note that on most of the properties, all models report accuracy $\ge 0.92$ even when trained on only 1\% of dataset. 
The results achieved demonstrate that relational properties are learnable, and that even simple ML models are effective in learning them.

\subsubsection{\RQTwo}

In \emph{RQ1} we show that ML models can exhibit high accuracy when learning relational properties.
Now, we move to the question of how well do those models generalize outside of the test set.
Our framework allows us to answer that question for decision tree models, which is focus of this section.
Specifically, we answer the question of how well the previously trained decision tree models perform in respect to the entire input space.
Table~\ref{tab:grdon_trainon} compares performance of the (same) decision trees models when evaluated: (1) on the test set, and (2) in respect to the entire input space.
Decision trees are trained using $10$\% of the constructed dataset (used in the previous section), while the test set represents the remaining part of the dataset.
Time shown is the total time taken by MCML for computing all $4$ performance metrics, where MCML leverages the model counting techniques to predict the number of true positives and negatives, and false positives and negatives.
To evaluate in respect to the entire input space we use the ground truth formula constrained with symmetry breaking conditions~\footnote{Symmetry breaking conditions are added so as to make distributions of examples similar to the ones present in the training set. We later show evaluation where we remove this constraint.}.

\begin{table}
  \Caption{Classification results on the test set for \emph{PartialOrder} property. ``\emph{Training:Test}'' ratio is shown in the left column, and symmetry breaking is turned off when creating datasets.}
  \label{tab:summarypartialoff}
	\centering
	\begin{footnotesize}
		\begin{tabular}{|@{\hspace{1ex}}r@{\hspace{1ex}}|@{\hspace{1ex}}c@{\hspace{1ex}}|@{\hspace{1ex}}c@{\hspace{1ex}}|@{\hspace{1ex}}c@{\hspace{1ex}}|@{\hspace{1ex}}c@{\hspace{1ex}}|@{\hspace{1ex}}c@{\hspace{1ex}}|}
			\hline
			\emph{Ratio} & \emph{Model}& \emph{Accuracy} & \emph{Precision} & \emph{Recall} & \emph{F1-score}\\
			\hline
		{ 75:25 } & DT & 0.9985 & 0.9970 & 1.0000 & 0.9985 \\
		& RFT & 0.9981 & 0.9963 & 1.0000 & 0.9981 \\
		& GBDT  & 0.9788 & 0.9593 & 1.0000 & 0.9792 \\
		& ABT &  0.8414 & 0.8636 & 0.8098 & 0.8359 \\
		& SVM &  0.9940 & 0.9881 & 1.0000 & 0.9940 \\
		& MLP &  0.9989 & 0.9978 & 1.0000 & 0.9989 \\
		\hline
		{ 25:75 } & DT & 0.9966 & 0.9935 & 0.9997 & 0.9966 \\
		& RFT &  0.9963 & 0.9927 & 1.0000 & 0.9963 \\
		& GBDT & 0.9780 & 0.9578 & 1.0000 & 0.9784 \\
		& ABT & 0.8394 & 0.8613 & 0.8085 & 0.8341\\ 
		& SVM &  0.9901 & 0.9806 & 1.0000 & 0.9902 \\
		& MLP &  0.9977 & 0.9954 & 1.0000 & 0.9977 \\ \hline
		{ 1:99 } & DT &  0.9692 & 0.9511 & 0.9893 & 0.9698\\ 
		& RFT &  0.9821 & 0.9699 & 0.9950 & 0.9823 \\
		& GBDT & 0.9734 & 0.9495 & 1.0000 & 0.9741 \\
		& ABT & 0.8352 & 0.8409 & 0.8268 & 0.8338\\ 
		& SVM &  0.9635 & 0.9320 & 1.0000 & 0.9648 \\
		& MLP & 0.9842 & 0.9704 & 0.9988 & 0.9844 \\
		
			\hline
		\end{tabular}
	\end{footnotesize}
\vspace{-5mm}	
\end{table}

\begin{table*}
  \centering
  \Caption{Decision tree performance on the test set and in respect to the entire state space encoded by ground truth formula $\phi$}
  \label{tab:grdoff_trainoff}
    \begin{footnotesize}
		\begin{tabular}{|@{\hspace{1ex}}l@{\hspace{1ex}}|@{\hspace{1ex}}r@{\hspace{1ex}}|@{\hspace{1ex}}r@{\hspace{1ex}}|@{\hspace{1ex}}c@{\hspace{1ex}}|@{\hspace{1ex}}c@{\hspace{1ex}}|@{\hspace{1ex}}r@{\hspace{1ex}}|@{\hspace{1ex}}c@{\hspace{1ex}}|@{\hspace{1ex}}c@{\hspace{1ex}}|@{\hspace{1ex}}c@{\hspace{1ex}}|@{\hspace{1ex}}c@{\hspace{1ex}}|}
			\hline
\emph{Property} & \emph{Accuracy} & \emph{Precision} & \emph{Recall} & \emph{F1-score}  & \emph{Accuracy} & \emph{Precision} & \emph{Recall} & \emph{F1-score}& \emph{Time[s]}\\ 
\emph{} & \emph{Test} & \emph{Test} & \emph{Test} & \emph{Test}  & \emph{$\phi$} & \emph{$\phi$} & \emph{$\phi$} & \emph{$\phi$}& \emph{}\\ 
\hline
\emph{Antisymmetric} &0.9997&0.9996&0.9998&0.9997      	& 	0.9996	& 	0.9935	& 	0.9998	& 	0.9967& 	2.1	\\ \hline
\emph{Bijective} 	 &0.9991&0.9982&1.0000&0.9991      	& 0.9981 & 0.0000 & 1.0000 & 0.0000&  225.8 \\ \hline
\emph{Connex}		 &0.9957&0.9933&0.9982&0.9957      	& 0.9935 & 0.2258 & 0.9985 & 0.3683&  0.8 \\ \hline
\emph{Equivalence}   &0.9997&0.9994&1.0000&0.9997      	& 0.9995 & 0.0000 & 1.0000 & 0.0000& 34.1 \\ \hline
\emph{Function} 	 &0.9946&0.9900&0.9993&0.9946      	& 0.9899 & 0.0001 & 0.9993 & 0.0001& 2.4 \\ \hline
\emph{Functional}	 &0.9968&0.9940&0.9997&0.9969      	& 0.9945 & 0.0003 & 0.9997 & 0.0006& 2.6 \\ \hline
\emph{Injective} 	 &0.9968&0.9940&0.9997&0.9969      	& 0.9877 & 0.0001 & 0.9989 & 0.0001& 2.3 \\ \hline
\emph{Irreflexive}	 &1.0000&1.0000&1.0000&1.0000     	& 1.0000 & 1.0000 & 1.0000 & 1.0000&0.5 \\ \hline
\emph{NonStrictOrder}&0.9990&0.9985&0.9994&0.9990     	& 0.9983 & 0.0011 & 0.9995 & 0.0022& 1.9 \\ \hline
\emph{PartialOrder}  &0.9934&0.9879&0.9991&0.9935     	& 0.9864 & 0.2407 & 0.9992 & 0.3879& 1.2 \\ \hline
\emph{PreOrder} 	 &0.9985&0.9974&0.9996&0.9985     	& 0.9972 & 0.0012 & 0.9997 & 0.0024& 2.0 \\ \hline
\emph{Reflexive} 	 &1.0000&1.0000&1.0000&1.0000   	&1.0000 & 1.0000 & 1.0000 & 1.0000& 0.5\\ \hline
\emph{StrictOrder} 	 &0.9988&0.9979&0.9997&0.9988    	& 0.9979 & 0.0009 & 0.9998 & 0.0019& 1.9 \\ \hline
\emph{Surjective} 	 &0.9988&0.9979&0.9997&0.9988     	& 0.9984 & 0.0000 & 1.0000 & 0.0000&  283.3 \\ \hline
\emph{TotalOrder}	 &0.9999&0.9997&1.0000&0.9999      	& 0.9997 & 	0.0000 & 	1.0000 & 	0.0000&  	8.4	 \\ \hline
\emph{Transitive} 	 &0.9999&0.9997&1.0000&0.9999   	& 0.9760 & 0.1588 & 0.9902 & 0.2737&  2.0 \\ \hline

	\end{tabular}       
	\end{footnotesize}
\end{table*}

When decision trees are evaluated on the test set, accuracy, recall and F1-score is $\ge 0.99$ while precision is $\ge0.98$.
However, when decision trees are evaluated on ground truth, minimum accuracy decreases to $0.89$, minimum precision and F1-score decreases to $0.00$ and minimum recall remains at $0.99$.
Precision and F1-score is spread all over the range 0~to~1, and is generally low.
In-fact, out of 15 properties that did not time-out, precision is around 0 on 12 properties.
This is because of high false positive rate showing that the decision trees are classifying many negative instances as positives.
This indicates that the models are biased towards classifying an example as positive, likely learning patterns present in training dataset but \emph{not generalizable}---applying the models outside of the dataset will likely incur many false positives.
For the properties Reflexive and Irreflexive, the models continue to have perfect performance since establishing these properties requires only checking the diagonal.
To summarize the findings, while the results in previous section showed encouraging results of learnability of relational properties, evaluation in respect to the entire input space shows concerning issues (with false positives) if models are to be used in the wild.

In summary, MCML's ability to quantify w.r.t. the entire state space is of unique value as it quantifies model's generalizability, and avoids a false sense of confidence in traditional ML metrics.
In addition, MCML is also time efficient as it reported results for 12 properties within 10 seconds and 3 properties within 220 seconds.
It timed-out on only one property (\emph{Equivalence}), where the state space for possible solutions is around $2^{400}$.     

\begin{table*}
  \centering
  \Caption{Decision tree performance on the test set (symmetries broken) and in respect to the entire state space encoded by ground truth formula $\phi$.}
  \label{tab:grdoff_trainon}
\begin{footnotesize}
\begin{tabular}{|@{\hspace{1ex}}l@{\hspace{1ex}}|@{\hspace{1ex}}r@{\hspace{1ex}}|@{\hspace{1ex}}r@{\hspace{1ex}}|@{\hspace{1ex}}c@{\hspace{1ex}}|@{\hspace{1ex}}c@{\hspace{1ex}}|@{\hspace{1ex}}r@{\hspace{1ex}}|@{\hspace{1ex}}c@{\hspace{1ex}}|@{\hspace{1ex}}c@{\hspace{1ex}}|@{\hspace{1ex}}c@{\hspace{1ex}}|@{\hspace{1ex}}c@{\hspace{1ex}}|}
\hline
\emph{Property} & \emph{Accuracy} & \emph{Precision} & \emph{Recall} & \emph{F1-score}  & \emph{Accuracy} & \emph{Precision} & \emph{Recall} & \emph{F1-score}& \emph{Time[s]}\\ 
\emph{} & \emph{Test} & \emph{Test} & \emph{Test} & \emph{Test}  & \emph{$\phi$} & \emph{$\phi$} & \emph{$\phi$} & \emph{$\phi$}& \emph{}\\ 
\hline
\emph{Antisymmetric}  &0.9913&0.9871&0.9957&0.9914      & 	0.9442 & 	0.5098 & 	0.2241 & 	0.3114&  1.3  \\ \hline
\emph{Bijective}      &0.9971&0.9942&1.0000&0.9971       & - & - & - & -&   - \\ \hline
\emph{Connex}    	  &0.9996&0.9993&1.0000&0.9996      & 0.9992 & 0.0520 & 0.1731 & 0.0800 &   1.9 \\ \hline
\emph{Equivalence}    &0.9958&0.9920&0.9997&0.9958      & - & - & - & - &  - \\ \hline
\emph{Function}       &0.9954&0.9928&0.9980&0.9954     	& 0.9922 & 0.0000 & 0.0667 & 0.0000&   86.6  \\ \hline
\emph{Functional}     &0.9991&0.9983&0.9998&0.9991      & 0.9980 & 0.0000 & 0.2907 & 0.0000&  277.5  \\ \hline
\emph{Injective}      &0.9979&0.9961&0.9998&0.9980     	& 0.9961 & 0.0000 & 0.2565 & 0.0000&  79.6  \\ \hline
\emph{Irreflexive}    &1.0000&1.0000&1.0000&1.0000     	&  1.0000 & 1.0000 & 1.0000 & 1.0000 &0.5 \\ \hline
\emph{NonStrictOrder} &0.9994&0.9989&1.0000&0.9994    	& 0.9990 & 0.0000 & 0.6263 & 0.0000&  8.9 \\ \hline
\emph{PartialOrder}   &0.9963&0.9936&0.9990&0.9963   	& 0.9937 & 0.0068 & 0.3435 & 0.0134 &  2.9 \\ \hline
\emph{PreOrder}       &0.9992&0.9985&0.9999&0.9992    	& 0.9987 & 0.0000 & 0.5180 & 0.0000&  16.6  \\ \hline
\emph{Reflexive}      &1.0000&1.0000&1.0000&1.0000    	& 1.0000 & 1.0000 & 1.0000 & 1.0000&   0.6  \\ \hline
\emph{StrictOrder}    &0.9991&0.9982&1.0000&0.9991    	& 0.9983 & 0.0000 & 0.4660 & 0.0000&  8.3 \\ \hline
\emph{Surjective}     &0.9980&0.9961&1.0000&0.9980   	&   - & - & - & - & - \\ \hline
\emph{TotalOrder}     &0.9994&0.9988&1.0000&0.9994    	& 	0.9990 & 	0.0000 & 	0.4737 & 	0.0000&  	3059.5  \\ \hline
\emph{Transitive}     &0.9949&0.9910&0.9989&0.9949     	& 0.9914 & 0.0038 & 0.2394 & 0.0074&   3.6 \\ \hline

\end{tabular}
\end{footnotesize}
\end{table*}

\vspace{-3mm}
\subsubsection{\RQThree}

We now move on to study the effect of symmetries in the dataset on ML models performance.
We perform experiments similar as in the \emph{RQ1} and \emph{RQ2} now shown in Table~\ref{tab:summarypartialoff} and Table~\ref{tab:grdoff_trainoff}, but without performing symmetry breaking on training and evaluation sets.

Table ~\ref{tab:summarypartialoff} summarizes performance of ML classifiers for one selected property (PartialOrder).
The dataset is generated without performing symmetry breaking.
Across all ratios, the accuracy is $\ge 0.83$, the precision is $\ge 0.84$, the recall is $\ge 0.80$, and the F1-score is $\ge 0.83$.
We have performed experiments for other properties (available at GitHub repository), and noticed similar trends.
Across all 16~properties, all 5~ratios, and all 6~ML models, the accuracy is $\ge 0.82$, the precision is $\ge 0.81$, the recall is $\ge 0.81$, and the F1-score is $\ge 0.82$.
Similar to before, all models achieve good performance, and good performance is preserved even for small training:test ratio.
However, in comparison to the previous results (Table~\ref{tab:summarypartialon}) there is a noticeable decrease in terms of accuracy and F1-score, introduced by symmetries.
This shows that if symmetries are broken (in both train and test set) than model can better learn the properties, as it only uses the representatives of distinct groups for learning.
This is analogous to training a digit classifier where all digits are upright.
Introducing symmetries, e.g., digits at different orientations makes it harder for classifier to learn well.

Table~\ref{tab:grdoff_trainoff} compares results of the decision tree models on the test set and in respect to the entire input space.
Unlike in \emph{RQ2}, now the training set is generated without symmetry breaking, and the entire input space encoded by the formula $\phi$ is now not constrained with symmetry breaking.
On the test set accuracy, recall and F1-score is $\ge0.99$ and precision is $\ge0.98$.
However, in respect to the entire input space, minimum accuracy is $0.98$, minimum precision and F1-score decreases to $0.00$, but minimum recall remains at $0.99$.
The outliers are properties Reflexive and Irreflexive, where the models continue to have perfect performance.
In summary, the results shows a similar trend as before, where even with enhancing the training set with symmetric examples, decision tree models still generalize poorly.
\vspace{-2mm}
\subsubsection{\RQFour}

\begin{table*}
	\centering
        \Caption{Decision tree performance on the test set and in respect to the entire state space encoded by ground truth formula $\phi$ (constrained by symmetry breaking).}
        \label{tab:grdon_trainoff}
	\begin{footnotesize}
		\begin{tabular}{|@{\hspace{1ex}}l@{\hspace{1ex}}|@{\hspace{1ex}}r@{\hspace{1ex}}|@{\hspace{1ex}}r@{\hspace{1ex}}|@{\hspace{1ex}}c@{\hspace{1ex}}|@{\hspace{1ex}}c@{\hspace{1ex}}|@{\hspace{1ex}}r@{\hspace{1ex}}|@{\hspace{1ex}}c@{\hspace{1ex}}|@{\hspace{1ex}}c@{\hspace{1ex}}|@{\hspace{1ex}}c@{\hspace{1ex}}|@{\hspace{1ex}}c@{\hspace{1ex}}|}
			\hline
\emph{Property} & \emph{Accuracy} & \emph{Precision} & \emph{Recall} & \emph{F1-score}  & \emph{Accuracy} & \emph{Precision} & \emph{Recall} & \emph{F1-score}& \emph{Time[s]}\\ 
\emph{} & \emph{Test} & \emph{Test} & \emph{Test} & \emph{Test}  & \emph{$\phi$} & \emph{$\phi$} & \emph{$\phi$} & \emph{$\phi$}& \emph{}\\ 
\hline
\emph{Antisymmetric}   &0.9997&0.9996&0.9998&0.9997    	& 	1.0000 & 	1.0000 & 	1.0000 & 	1.0000& 0.7 \\ \hline
	\emph{Bijective}   &0.9991&0.9982&1.0000&0.9991    	& 0.9992 & 0.0000 & 1.0000 & 0.0000& 10.1 \\ \hline
	\emph{Connex}      &0.9957&0.9933&0.9982&0.9957    	& 0.9959 & 0.1510 & 1.0000 & 0.2624& 0.6 \\ \hline
	\emph{Equivalence} &0.9997&0.9994&1.0000&0.9997    	& 0.9991 & 0.0000 & 1.0000 & 0.0000& 15.7 \\ \hline
	\emph{Function}    &0.9946&0.9900&0.9993&0.9946    	& 0.9922 & 0.0000 & 0.9912 & 0.0001& 1.6 \\ \hline
	\emph{Functional}  &0.9968&0.9940&0.9997&0.9969    	& 0.9964 & 0.0003 & 1.0000 & 0.0005& 1.5 \\ \hline
	\emph{Injective}   &0.9968&0.9940&0.9997&0.9969  	& 0.9914 & 0.0002 & 1.0000 & 0.0003& 1.5 \\ \hline
	\emph{Irreflexive} &1.0000&1.0000&1.0000&1.0000   	& 1.0000 & 1.0000 & 1.0000 & 1.0000& 0.4 \\ \hline
	\emph{NonStrictOrder}&0.9990&0.9985&0.9994&0.9990  	& 0.9972 & 0.0016 & 1.0000 & 0.0032&  1.5 \\ \hline
	\emph{PartialPrder}&0.9934&0.9879&0.9991&0.9935    	& 0.9862 & 0.2784 & 1.0000 & 0.4356& 1.0 \\ \hline
	\emph{PreOrder}    &0.9985&0.9974&0.9996&0.9985    	& 0.9971 & 0.0028 & 0.9997 & 0.0056& 1.6 \\ \hline
	\emph{Reflexive}   &1.0000&1.0000&1.0000&1.0000    	& 1.0000 & 1.0000 & 1.0000 & 1.0000& 0.5 \\ \hline
	\emph{StrictOrder} &0.9988&0.9979&0.9997&0.9988    	& 0.9977 & 0.0021 & 1.0000 & 0.0041& 1.5 \\ \hline
	\emph{Surjective}  &0.9988&0.9979&0.9997&0.9988     & 0.9968 & 0.0000 & 1.0000 & 0.0000& 10.9 \\ \hline
	\emph{TotalOrder}  &0.9999&0.9997&1.0000&0.9999     & 	0.9996 & 	0.0000 & 	1.0000 & 	0.0000&  	9.4 \\ \hline
	\emph{Transitive}  &0.9999&0.9997&1.0000&0.9999    	& 0.9906 & 0.1673 & 0.9912 & 0.2863& 0.8 \\ \hline

		\end{tabular}
	\end{footnotesize}
	
\end{table*}

\begin{table*}
\centering
\Caption{Evaluating differences between decision tree models.}
\label{tab:singledtcomparisononb}
\begin{footnotesize}
		\begin{tabular}{|@{\hspace{1ex}}l@{\hspace{1ex}}|@{\hspace{1ex}}c@{\hspace{1ex}}|@{\hspace{1ex}}c@{\hspace{1ex}}|@{\hspace{1ex}}r@{\hspace{1ex}}|@{\hspace{1ex}}c@{\hspace{1ex}}|@{\hspace{1ex}}c@{\hspace{1ex}}|@{\hspace{1ex}}c@{\hspace{1ex}}|}
\hline
\emph{Subject}  & \emph{TT} & \emph{TF} & \emph{FT} & \emph{FF}& \emph{Diff}& \emph{Time [s]} \\
\hline
\emph{Antisymmetric}&7.86E+05&3.28E+04&3.17E+04&3.36E+07&0.19&1.1\\ \hline
\emph{Bijective}&9.19E+54&5.88E+56&3.86E+56&1.00E+59&0.96&117.8\\ \hline
\emph{Connex}&3.62E+07&1.15E+07&1.21E+07&6.87E+10&0.03&1.7\\ \hline
\emph{Equivalence}&7.88E+116&1.95E+118&3.91E+118&2.54E+120&2.25&852.8\\ \hline
\emph{Function}&6.31E+16&8.11E+16&8.11E+16&1.84E+19&0.87&8.1\\ \hline
\emph{Functional}&2.03E+16&1.58E+16&1.58E+16&1.84E+19&0.17&7.4\\ \hline
\emph{Injective}&7.21E+16&0&0&1.84E+19&0.00&4.2\\ \hline
\emph{Irreflexive}&1.05E+06&0&0&3.25E+07&0.00&0.5\\ \hline
\emph{NonStrictOrder}&2.92E+11&2.92E+11&2.92E+11&5.63E+14&0.10&4\\ \hline
\emph{PartialOrder}&3.02E+08&1.07E+08&1.17E+08&6.87E+10&0.32&2.4\\ \hline
\emph{PreOrder}&3.78E+11&3.78E+11&3.78E+11&5.63E+14&0.13&3.8\\ \hline
\emph{Reflexive}&1.05E+06&0&0&3.25E+07&0.00&0.5\\ \hline
\emph{StrictOrder}&6.18E+11&3.44E+11&3.44E+11&5.63E+14&0.12&3.8\\ \hline
\emph{Surjective}&2.45E+55&3.68E+56&3.68E+56&1.00E+59&0.73&119.2\\ \hline
\emph{TotalOrder}&3.65E+47&3.65E+47&3.65E+47&7.48E+50&0.10&79.5\\ \hline
\emph{Transitive}&4.36E+08&2.01E+08&1.93E+08&6.87E+10&0.57&2.4\\ \hline

\end{tabular}
\end{footnotesize}
\end{table*}

We now move to the next question of how the mismatch in presence of symmetries in training and evaluation sets affect the performance.
Specifically, we look at the two scenarios: (1) symmetries are not present in the training set but are in the evaluation set, and (2) symmetries are present in the training set but not in the evaluation.

Table~\ref{tab:grdoff_trainon} shows results of decision tree models trained and tested on datasets with symmetry breaking, but evaluated on the entire input space (without symmetry breaking).
When decision trees are evaluated on the test set accuracy, recall and F1-score is $\ge0.99$, while precision is $\ge0.98$.
However, when decision trees are evaluated on ground truth (without symmetry breaking constraints), minimum accuracy decrease to $0.94$, minimum precision and F1-score decreases to $0.00$, while minimum recall decrease to $0.06$.
Since symmetries are present in the entire input space, the performance decreases compared to the case where symmetries are also present in the training set (Table~\ref{tab:grdoff_trainoff})
These experiments show that decision trees perform worst when they are trained on the dataset without symmetries and evaluated w.r.t entire state space (with no symmetry breaking).
This is expected result since the trained model did not see any symmetrical instances in the training dataset and is likely to perform incorrectly when tested on permutations of same instances.
This is similar to an example of neural networks failing to generalize on digit recognition task, when seeing digits at different orientation than observed during the training~\cite{nnLearnInvariants}.

Table~\ref{tab:grdon_trainoff} shows results of decision tree models trained and tested on datasets without symmetry breaking, and evaluated on the entire input space with added symmetry breaking constraints.
When decision trees are evaluated on the test set, accuracy, recall and F1-score is $\ge0.99$ while precision is $\ge0.98$.
However, when decision trees are evaluated on ground truth (with added symmetry breaking), minimum accuracy remains at $0.99$, minimum precision and F1-score decreases to $0.00$ while minimum recall remains at $0.99$.
These results show that even when training set is richer (contains symmetries) than evaluation set, ML models still fail to generalize in respect to the entire input space.
\vspace{-3mm}
\\
\subsubsection{\RQFive}

We next employ MCML to quantify the difference between the two decision tree models.
Our framework allows us to get rigorous measure of the differences between the two models, i.e., the measure in respect to the entire input space (not just the train/test datasets).
We trained two decision tree models for each property, using different values of hyperparameters, and measured difference between the two.

Table~\ref{tab:singledtcomparisononb} shows the results for quantifying differences between 2~trained models, where we show number of examples in which both models predict true (TT), in which the first predicts true and second false (TF), and the other combinations denoted by FT and FF.
The \emph{Diff} column shows the percentage of cases in which the two models make a different prediction.
In all cases, the difference is close to~$0$. MCML is able to quantify the differences between the two models in respect to the entire input space, which makes it a powerful technique for evaluation of models. For example, one can take a smaller (compressed) model and rigorously quantify it in respect to the larger model to see if it can be used as a replacement. MCML is able to detect this for 12 properties within 10 seconds and is able to detect all properties within 1000 seconds. 

\subsection{Discussion}

\subsubsection{Traditional Metrics and MCML}
Tables~\ref{tab:summarypartialon} and~\ref{tab:summarypartialoff}
evaluate off-the-shelf models using traditional ML metrics based on
training and test datasets.  These metrics give a false sense of
confidence, which our proposed MCML framework addresses.
Tables~\ref{tab:grdon_trainon},~\ref{tab:grdoff_trainoff},~\ref{tab:grdoff_trainon}
and~\ref{tab:grdon_trainoff} show that the majority of precision
scores and F1-scores are low w.r.t the entire state space.  An
important characteristic of the properties we consider is that the
number of positive cases is far too small compared to the number of
negative cases.  We believe new classifiers are needed to handle
complex relational properties.

Alloy formulas are intuitively simple but semantically quite complex,
and getting high precision with respect to the entire state space is
hard.  In fact, in only 2 cases (reflexive and irreflexive) the
precision is 1, and that is because the properties can be checked
simply by inspecting the diagonal entries -- the trained decision
trees indeed do so. In all other cases, the precision is very low
(<0.1).  MCML provides a new tool to rigorously study properties that
seem ``easy'' to learn in the traditional setting of training and test
data, but are actually ``hard'' to learn when viewed in the context of
the entire state space.  For example, for partial order, precision was
0.9936 in traditional setting whereas MCML reported that the precision
is 0.0059 for the entire state space.  This indicates that the model is
biased towards classifying an example as a partial order, likely
learning patterns present in training dataset but not generalizable --
when using data outside of dataset one can anticipate many false
positives.  MCML's ability to quantify w.r.t. the entire state space
is of unique value as it quantifies model's generalizability, and
avoids a false sense of confidence in traditional ML metrics.

\begin{table}
  \Caption{Comparison of performance between traditional metrics and
    MCML for \emph{antisymmetric} property for different class ratios
    (ratio of valid samples to invalid samples in training dataset);
    true ratio for the entire state space is 1:99.}
    \label{tab:rebuttal}
	\centering
	\begin{footnotesize}
		\begin{tabular}{|C{2.0cm}|C{2.0cm}|C{2.0cm}|}
			\hline
			\emph{Ratio of Valid:Invalid in training dataset} & \emph{Traditional Precision}& \emph{MCML Precision}\\
			\hline
		{ 99:1 } & 0.98 & 0.19 \\ \hline
		{ 90:10 } & 0.97 & 0.21 \\ \hline
		{ 75:25 } & 0.98 & 0.44 \\ \hline
		{ 50:50 } & 0.99 & 0.46 \\ \hline
		{ 25:75 } & 0.99 & 0.56 \\ \hline
		{ 10:90 } & 0.99 & 0.75 \\ \hline
		{  1:99 } & 1.00 & 0.97 \\
			\hline
		\end{tabular}
	\end{footnotesize}
	
\end{table}

\noindent\textbf{Varying class ratios (ratio of valid samples to
    invalid samples in training dataset).}  A key utility of MCML is
  that it allows quantifying model generalizability even if the
  training distribution is different from the true distribution.
  Table~\ref{tab:rebuttal} illustrates this for the antisymmetric
  property.  The traditional metric reports precision to be $\ge$0.97
  for all ratios whereas MCML metrics report that the precision is as
  low as 0.19 (for 99:1 ratio) and is $\ge$0.97 only when the models
  are trained on datasets having a class ratio of 1:99, which is very
  close to the true distribution ratio.  Thus, for this property,
  traditional metrics fail to capture the performance for almost all
  class ratios; in contrast, MCML metrics allow precise quantification
  of true performance of the trained model for each class ratio.

The primary limitation of MCML is its requirement for $\phi$, which
characterizes the ground truth and is not required by traditional
metrics (which are not able to utilize it).  We expect property
synthesis methods can help MCML apply more generally; moreover, even
when the exact $\phi$ is not available, an approximation of $\phi$
that is human understandable can be of much use.

\subsubsection{Use of Alloy/SAT}
  We used the Alloy toolset for writing the properties and
  generating the datasets for this study.  The use of Alloy introduces
  a potential for bias in the results due to the specific techniques
  employed by Alloy.  Our study mitigates this potential threat to
  validity as follows.  There are two sources of potential bias:
  1)~translation from Alloy to propositional logic; and 2)~solving the
  propositional formula.  When Alloy's symmetry breaking is turned
  off, the propositional formula Alloy creates faithfully represents
  the original property (with respect to the scope).  Specifically,
  the set of solutions at the Alloy level is the same (modulo data
  representation) as the set of solutions at the propositional level
  with respect to the \emph{primary} variables.  For positive data
  samples, recall, we use every solution enumerated by the SAT
  backend.  Any SAT solver that enumerates every solution will create
  the same set of solutions.  Indeed, different solvers may create
  that set in different orders.  However, our experiments do not
  depend on the order in which the solutions are created.  For
  example, when we use 10\% of valid solutions for training, we do not
  select the first 10\% of the solutions created by the solver, rather
  we select a random subset with the desired size from all the
  solutions.  For negative data samples, recall, we do not use any
  constraint solving at all, and instead select a set of negative
  solutions at random (as described earlier in this section).
  Hence, when symmetry breaking is turned off, our results do not
  suffer from bias from the use of Alloy.

  When symmetry breaking is turned on, Alloy adds symmetry breaking
  predicates to the propositional formula before it is solved by the
  backend SAT engine.  Alloy's motivation of adding these predicates,
  which preserve the satisfiability of the original formula, is to
  enable faster solving (by helping SAT prune more effectively,
  especially for formulas that are unsatisfiable).  A consequence of
  adding them for our study is that they remove \emph{valid} solutions
  and can substantially reduce the number of solutions, although the
  actual reduction depends on the specific formula and scope.  To
  illustrate, consider the common example of enumerating binary
  trees~\cite{CLRS}.  For trees with 3 nodes, Alloy removes all
  symmetries and gives a 6X reduction in the number of solutions; but
  for 7 nodes, it gives a 1160X reduction whereas full symmetry
  breaking gives 5040X reduction~\cite{StudySymmetry}.  Thus, under
  symmetry breaking turned on, our results are specific to the
  symmetry breaking predicates added by Alloy's default setting for
  each property and scope.  A different setting in Alloy or a
  different tool may lead to different results.  We plan to more
  deeply study the impact of symmetry breaking on learnability in
  future work.

Note also that MCML uses a very efficient encoding of decision trees
directly into CNF that does not use any \emph{auxiliary} variables
and is linear in the size of the tree.  Thus, the translation is not
only faithful but also allows the resulting formulas to be solved
readily by off-the-shelf model counters.


	\section{Conclusion}

This paper introduced the MCML approach for empirically studying the
learnability of relational properties that can be expressed in the
software design language Alloy.  A key novelty of MCML is
quantification of the performance of and semantic differences among
trained machine learning (ML) models, specifically decision trees,
with respect to entire input spaces (up to a bound on the input size),
and not just for given training and test datasets (as is the common
practice).  MCML reduces the quantification problems to the classic
complexity theory problem of model counting.  The results show that relatively simple ML models can
achieve surprisingly high performance (accuracy and F1-score)
when evaluated in the common setting of
using training and test datasets, indicating the seeming
simplicity of learning these properties.  However, the use of MCML
metrics shows that the performance can degrade
substantially when tested against the entire (bounded) input space,
indicating the high complexity of precisely learning these properties,
and the usefulness of model counting in quantifying the true performance.

MCML offers exciting new directions for leveraging model counting in
machine learning, e.g., 1) it provides quantitative answers to key
questions like "did we train enough?", "how much did we overfit?", and
"is this model basically the same as this other model (I have in
mind)?"; and 2) it allows informed decision making for ML-based
systems, e.g., "should a deployed model be replaced with another
(newer) model?".


\begin{acks}
We thank Darko Marinov and the reviewers for very helpful comments and feedback.
This work was partially supported by the National Science Foundation
grant CCF-1718903.
\end{acks}

\balance
	\clearpage
	\bibliography{bib.bib}


\begin{thebibliography}{73}


\ifx \showCODEN    \undefined \def \showCODEN     #1{\unskip}     \fi
\ifx \showDOI      \undefined \def \showDOI       #1{#1}\fi
\ifx \showISBNx    \undefined \def \showISBNx     #1{\unskip}     \fi
\ifx \showISBNxiii \undefined \def \showISBNxiii  #1{\unskip}     \fi
\ifx \showISSN     \undefined \def \showISSN      #1{\unskip}     \fi
\ifx \showLCCN     \undefined \def \showLCCN      #1{\unskip}     \fi
\ifx \shownote     \undefined \def \shownote      #1{#1}          \fi
\ifx \showarticletitle \undefined \def \showarticletitle #1{#1}   \fi
\ifx \showURL      \undefined \def \showURL       {\relax}        \fi
\providecommand\bibfield[2]{#2}
\providecommand\bibinfo[2]{#2}
\providecommand\natexlab[1]{#1}
\providecommand\showeprint[2][]{arXiv:#2}

\bibitem[\protect\citeauthoryear{Bagheri, Kang, Malek, and Jackson}{Bagheri
  et~al\mbox{.}}{2018}]%
        {BagheriETAL2018}
\bibfield{author}{\bibinfo{person}{Hamid Bagheri}, \bibinfo{person}{Eunsuk
  Kang}, \bibinfo{person}{Sam Malek}, {and} \bibinfo{person}{Daniel Jackson}.}
  \bibinfo{year}{2018}\natexlab{}.
\newblock \showarticletitle{A formal approach for detection of security flaws
  in the {Android} permission system}.
\newblock \bibinfo{journal}{\emph{Formal Asp. Comput.}} (\bibinfo{year}{2018}).
\newblock


\bibitem[\protect\citeauthoryear{Baluta, Shen, Shinde, Meel, and Saxena}{Baluta
  et~al\mbox{.}}{2019}]%
        {Baluta2019quantitative}
\bibfield{author}{\bibinfo{person}{Teodora Baluta}, \bibinfo{person}{Shiqi
  Shen}, \bibinfo{person}{Shweta Shinde}, \bibinfo{person}{Kuldeep~S Meel},
  {and} \bibinfo{person}{Prateek Saxena}.} \bibinfo{year}{2019}\natexlab{}.
\newblock \showarticletitle{Quantitative Verification of Neural Networks And
  its Security Applications}.
\newblock \bibinfo{journal}{\emph{arXiv preprint arXiv:1906.10395}}
  (\bibinfo{year}{2019}).
\newblock


\bibitem[\protect\citeauthoryear{Bastani, Pu, and Solar-Lezama}{Bastani
  et~al\mbox{.}}{2018}]%
        {BastaniETAL18VerifiableRL}
\bibfield{author}{\bibinfo{person}{Osbert Bastani}, \bibinfo{person}{Yewen Pu},
  {and} \bibinfo{person}{Armando Solar-Lezama}.}
  \bibinfo{year}{2018}\natexlab{}.
\newblock \showarticletitle{Verifiable reinforcement learning via policy
  extraction}. In \bibinfo{booktitle}{\emph{ASIACCS}}.
\newblock


\bibitem[\protect\citeauthoryear{Blumer, Ehrenfeucht, Haussler, and
  Warmuth}{Blumer et~al\mbox{.}}{1989}]%
        {Blumer:1989:LVD:76359.76371}
\bibfield{author}{\bibinfo{person}{Anselm Blumer}, \bibinfo{person}{A.
  Ehrenfeucht}, \bibinfo{person}{David Haussler}, {and}
  \bibinfo{person}{Manfred~K. Warmuth}.} \bibinfo{year}{1989}\natexlab{}.
\newblock \showarticletitle{Learnability and the {Vapnik}-{Chervonenkis}
  Dimension}.
\newblock \bibinfo{journal}{\emph{JACM}} \bibinfo{volume}{36},
  \bibinfo{number}{4} (\bibinfo{year}{1989}).
\newblock


\bibitem[\protect\citeauthoryear{Boyapati, Khurshid, and Marinov}{Boyapati
  et~al\mbox{.}}{2002}]%
        {Boyapati:2002:KAT:566172.566191}
\bibfield{author}{\bibinfo{person}{Chandrasekhar Boyapati},
  \bibinfo{person}{Sarfraz Khurshid}, {and} \bibinfo{person}{Darko Marinov}.}
  \bibinfo{year}{2002}\natexlab{}.
\newblock \showarticletitle{Korat: {Automated Testing Based on Java
  Predicates}}. In \bibinfo{booktitle}{\emph{ISSTA}}.
\newblock


\bibitem[\protect\citeauthoryear{{Brun} and {Ernst}}{{Brun} and
  {Ernst}}{2004}]%
        {1317470}
\bibfield{author}{\bibinfo{person}{Y. {Brun}} {and} \bibinfo{person}{M.~D.
  {Ernst}}.} \bibinfo{year}{2004}\natexlab{}.
\newblock \showarticletitle{Finding latent code errors via machine learning
  over program executions}. In \bibinfo{booktitle}{\emph{ICSE}}.
\newblock
\showISSN{0270-5257}


\bibitem[\protect\citeauthoryear{Can Deep Networks Learn Invariants}{Can Deep
  Networks Learn Invariants}{[n. d.]}]%
        {nnLearnInvariants}
Can Deep Networks Learn Invariants \bibinfo{year}{[n. d.]}\natexlab{}.
\newblock \bibinfo{title}{Can Deep Networks Learn Invariants}.
\newblock   (\bibinfo{year}{[n. d.]}).
\newblock
\newblock
\shownote{\url{https://blog.singularitynet.io/can-deep-networks-learn-invariants-1e06a5052555}.}


\bibitem[\protect\citeauthoryear{Chan and Darwiche}{Chan and Darwiche}{2003}]%
        {ChanDarwiche2003}
\bibfield{author}{\bibinfo{person}{Hei Chan} {and} \bibinfo{person}{Adnan
  Darwiche}.} \bibinfo{year}{2003}\natexlab{}.
\newblock \showarticletitle{Reasoning About Bayesian Network Classifiers}. In
  \bibinfo{booktitle}{\emph{UAI}}.
\newblock


\bibitem[\protect\citeauthoryear{Chavira and Darwiche}{Chavira and
  Darwiche}{2008}]%
        {Chavira2008OnPI}
\bibfield{author}{\bibinfo{person}{Mark Chavira} {and} \bibinfo{person}{Adnan
  Darwiche}.} \bibinfo{year}{2008}\natexlab{}.
\newblock \showarticletitle{On probabilistic inference by weighted model
  counting}.
\newblock \bibinfo{journal}{\emph{JAI}} \bibinfo{volume}{172},
  \bibinfo{number}{6-7} (\bibinfo{year}{2008}).
\newblock


\bibitem[\protect\citeauthoryear{{CheckMate GitHub}}{{CheckMate
  GitHub}}{2019}]%
        {CheckMateGitHub}
\bibfield{author}{\bibinfo{person}{{CheckMate GitHub}}.}
  \bibinfo{year}{2019}\natexlab{}.
\newblock \bibinfo{title}{\url{https://github.com/ctrippel/checkmate}}.
  (\bibinfo{year}{2019}).
\newblock


\bibitem[\protect\citeauthoryear{Chong, Sorensen, and Wickerson}{Chong
  et~al\mbox{.}}{2018}]%
        {ChongETAL2018}
\bibfield{author}{\bibinfo{person}{Nathan Chong}, \bibinfo{person}{Tyler
  Sorensen}, {and} \bibinfo{person}{John Wickerson}.}
  \bibinfo{year}{2018}\natexlab{}.
\newblock \showarticletitle{The Semantics of Transactions and Weak Memory in
  {x86}, {Power}, {ARM}, and {C++}}. In \bibinfo{booktitle}{\emph{PLDI}}.
\newblock


\bibitem[\protect\citeauthoryear{Cormen, Leiserson, Rivest, and Stein}{Cormen
  et~al\mbox{.}}{2009}]%
        {CLRS}
\bibfield{author}{\bibinfo{person}{Thomas~H. Cormen},
  \bibinfo{person}{Charles~E. Leiserson}, \bibinfo{person}{Ronald~L. Rivest},
  {and} \bibinfo{person}{Clifford Stein}.} \bibinfo{year}{2009}\natexlab{}.
\newblock \bibinfo{booktitle}{\emph{Introduction to Algorithms, Third Edition}
  (\bibinfo{edition}{3rd} ed.)}.
\newblock \bibinfo{publisher}{The MIT Press}.
\newblock
\showISBNx{0262033844}


\bibitem[\protect\citeauthoryear{de~Moura, Kong, Avigad, van Doorn, and von
  Raumer}{de~Moura et~al\mbox{.}}{2015}]%
        {MouraKADR15}
\bibfield{author}{\bibinfo{person}{Leonardo~Mendon{\c{c}}a de Moura},
  \bibinfo{person}{Soonho Kong}, \bibinfo{person}{Jeremy Avigad},
  \bibinfo{person}{Floris van Doorn}, {and} \bibinfo{person}{Jakob von
  Raumer}.} \bibinfo{year}{2015}\natexlab{}.
\newblock \showarticletitle{The {Lean} Theorem Prover (System Description)}. In
  \bibinfo{booktitle}{\emph{{CADE}}}.
\newblock


\bibitem[\protect\citeauthoryear{Demsky and Rinard}{Demsky and Rinard}{2003}]%
        {DBLP:conf/oopsla/DemskyR03}
\bibfield{author}{\bibinfo{person}{Brian Demsky} {and}
  \bibinfo{person}{Martin~C. Rinard}.} \bibinfo{year}{2003}\natexlab{}.
\newblock \showarticletitle{Automatic detection and repair of errors in data
  structures}. In \bibinfo{booktitle}{\emph{OOPSLA}}.
\newblock


\bibitem[\protect\citeauthoryear{E{\'e}n and S{\"o}rensson}{E{\'e}n and
  S{\"o}rensson}{2004}]%
        {MiniSAT2004}
\bibfield{author}{\bibinfo{person}{Niklas E{\'e}n} {and}
  \bibinfo{person}{Niklas S{\"o}rensson}.} \bibinfo{year}{2004}\natexlab{}.
\newblock \showarticletitle{An Extensible {SAT}-solver}. In
  \bibinfo{booktitle}{\emph{Theory and Applications of Satisfiability
  Testing}}, \bibfield{editor}{\bibinfo{person}{Enrico Giunchiglia} {and}
  \bibinfo{person}{Armando Tacchella}} (Eds.). \bibinfo{pages}{502--518}.
\newblock


\bibitem[\protect\citeauthoryear{Fierens, den Broeck, Thon, Gutmann, and
  Raedt}{Fierens et~al\mbox{.}}{2012}]%
        {FierensETAL2012}
\bibfield{author}{\bibinfo{person}{Daan Fierens}, \bibinfo{person}{Guy~Van den
  Broeck}, \bibinfo{person}{Ingo Thon}, \bibinfo{person}{Bernd Gutmann}, {and}
  \bibinfo{person}{Luc~De Raedt}.} \bibinfo{year}{2012}\natexlab{}.
\newblock \showarticletitle{Inference in Probabilistic Logic Programs using
  Weighted CNF's}.
\newblock \bibinfo{journal}{\emph{CoRR}}  \bibinfo{volume}{abs/1202.3719}
  (\bibinfo{year}{2012}).
\newblock


\bibitem[\protect\citeauthoryear{Galeotti, Rosner, Pombo, and Frias}{Galeotti
  et~al\mbox{.}}{2013}]%
        {TACOGaleottiETALTSE2013}
\bibfield{author}{\bibinfo{person}{J.~P. Galeotti}, \bibinfo{person}{N.
  Rosner}, \bibinfo{person}{C.~G.~L\'{o}pez Pombo}, {and}
  \bibinfo{person}{M.~F. Frias}.} \bibinfo{year}{2013}\natexlab{}.
\newblock \showarticletitle{{TACO}: Efficient {SAT}-Based Bounded Verification
  Using Symmetry Breaking and Tight Bounds}.
\newblock \bibinfo{journal}{\emph{TSE}} (\bibinfo{year}{2013}).
\newblock


\bibitem[\protect\citeauthoryear{García and Herrera}{García and
  Herrera}{2009}]%
        {doi:10.1162/evco.2009.17.3.275}
\bibfield{author}{\bibinfo{person}{Salvador García} {and}
  \bibinfo{person}{Francisco Herrera}.} \bibinfo{year}{2009}\natexlab{}.
\newblock \showarticletitle{Evolutionary Undersampling for Classification with
  Imbalanced Datasets: Proposals and Taxonomy}.
\newblock \bibinfo{journal}{\emph{Evolutionary Computation}}
  \bibinfo{volume}{17}, \bibinfo{number}{3} (\bibinfo{year}{2009}).
\newblock


\bibitem[\protect\citeauthoryear{Garg, Neider, Madhusudan, and Roth}{Garg
  et~al\mbox{.}}{2016}]%
        {Garg:2016:LIU:2837614.2837664}
\bibfield{author}{\bibinfo{person}{Pranav Garg}, \bibinfo{person}{Daniel
  Neider}, \bibinfo{person}{P. Madhusudan}, {and} \bibinfo{person}{Dan Roth}.}
  \bibinfo{year}{2016}\natexlab{}.
\newblock \showarticletitle{Learning Invariants Using Decision Trees and
  Implication Counterexamples}. In \bibinfo{booktitle}{\emph{POPL}}.
\newblock


\bibitem[\protect\citeauthoryear{Gens and Domingos}{Gens and Domingos}{2013}]%
        {GensDomingos2013}
\bibfield{author}{\bibinfo{person}{Robert Gens} {and} \bibinfo{person}{Pedro
  Domingos}.} \bibinfo{year}{2013}\natexlab{}.
\newblock \showarticletitle{Learning the Structure of Sum-product Networks}. In
  \bibinfo{booktitle}{\emph{ICML}}.
\newblock


\bibitem[\protect\citeauthoryear{Gomes, Sabharwal, and Selman}{Gomes
  et~al\mbox{.}}{2008}]%
        {Gomes08modelcounting}
\bibfield{author}{\bibinfo{person}{Carla~P. Gomes}, \bibinfo{person}{Ashish
  Sabharwal}, {and} \bibinfo{person}{Bart Selman}.}
  \bibinfo{year}{2008}\natexlab{}.
\newblock \bibinfo{title}{Model Counting}.
\newblock   (\bibinfo{year}{2008}).
\newblock


\bibitem[\protect\citeauthoryear{Gopinath, Khurshid, Saha, and
  Chandra}{Gopinath et~al\mbox{.}}{2014}]%
        {GopinathETALICSE2014}
\bibfield{author}{\bibinfo{person}{Divya Gopinath}, \bibinfo{person}{Sarfraz
  Khurshid}, \bibinfo{person}{Diptikalyan Saha}, {and} \bibinfo{person}{Satish
  Chandra}.} \bibinfo{year}{2014}\natexlab{}.
\newblock \showarticletitle{Data-guided repair of selection statements}. In
  \bibinfo{booktitle}{\emph{36th International Conference on Software
  Engineering (ICSE)}}. \bibinfo{pages}{243--253}.
\newblock


\bibitem[\protect\citeauthoryear{Gopinath, Malik, and Khurshid}{Gopinath
  et~al\mbox{.}}{2011}]%
        {GopinathETALTACAS2011}
\bibfield{author}{\bibinfo{person}{Divya Gopinath},
  \bibinfo{person}{Muhammad~Zubair Malik}, {and} \bibinfo{person}{Sarfraz
  Khurshid}.} \bibinfo{year}{2011}\natexlab{}.
\newblock \showarticletitle{Specification-Based Program Repair Using {SAT}}. In
  \bibinfo{booktitle}{\emph{{TACAS}}}.
\newblock


\bibitem[\protect\citeauthoryear{Gopinath, Wang, Zhang, Pasareanu, and
  Khurshid}{Gopinath et~al\mbox{.}}{2018}]%
        {GopinathETAL2018Arxiv}
\bibfield{author}{\bibinfo{person}{Divya Gopinath}, \bibinfo{person}{Kaiyuan
  Wang}, \bibinfo{person}{Mengshi Zhang}, \bibinfo{person}{Corina~S.
  Pasareanu}, {and} \bibinfo{person}{Sarfraz Khurshid}.}
  \bibinfo{year}{2018}\natexlab{}.
\newblock \showarticletitle{Symbolic Execution for Deep Neural Networks}.
\newblock \bibinfo{journal}{\emph{CoRR}}  \bibinfo{volume}{abs/1807.10439}
  (\bibinfo{year}{2018}).
\newblock


\bibitem[\protect\citeauthoryear{H{\aa}stad}{H{\aa}stad}{1987}]%
        {article1987}
\bibfield{author}{\bibinfo{person}{Johan H{\aa}stad}.}
  \bibinfo{year}{1987}\natexlab{}.
\newblock \bibinfo{booktitle}{\emph{Computational Limitations of Small-depth
  Circuits}}.
\newblock \bibinfo{publisher}{MIT Press}, \bibinfo{address}{Cambridge, MA,
  USA}.
\newblock
\showISBNx{0262081679}


\bibitem[\protect\citeauthoryear{Heule, J\''{a}rvisalo, and Suda}{Heule
  et~al\mbox{.}}{2018}]%
        {ProcedingsSAT2018}
\bibfield{author}{\bibinfo{person}{Marijn J.~H. Heule},
  \bibinfo{person}{Matti~Juhani J\''{a}rvisalo}, {and} \bibinfo{person}{Martin
  Suda}.} \bibinfo{year}{2018}\natexlab{}.
\newblock \showarticletitle{Proceedings of {SAT} Competition 2018: Solver and
  Benchmark Descriptions} \emph{(\bibinfo{series}{Department of Computer
  Science Series of Publications B})}.
\newblock
\urldef\tempurl%
\url{http://hdl.handle.net/10138/237063}
\showURL{%
\tempurl}


\bibitem[\protect\citeauthoryear{Huang, Kwiatkowska, Wang, and Wu}{Huang
  et~al\mbox{.}}{2017}]%
        {HuangKWW17}
\bibfield{author}{\bibinfo{person}{Xiaowei Huang}, \bibinfo{person}{Marta
  Kwiatkowska}, \bibinfo{person}{Sen Wang}, {and} \bibinfo{person}{Min Wu}.}
  \bibinfo{year}{2017}\natexlab{}.
\newblock \showarticletitle{Safety Verification of Deep Neural Networks}. In
  \bibinfo{booktitle}{\emph{CAV}}.
\newblock


\bibitem[\protect\citeauthoryear{Hubara, Courbariaux, Soudry, El-Yaniv, and
  Bengio}{Hubara et~al\mbox{.}}{2016}]%
        {hubara2016binarized}
\bibfield{author}{\bibinfo{person}{Itay Hubara}, \bibinfo{person}{Matthieu
  Courbariaux}, \bibinfo{person}{Daniel Soudry}, \bibinfo{person}{Ran
  El-Yaniv}, {and} \bibinfo{person}{Yoshua Bengio}.}
  \bibinfo{year}{2016}\natexlab{}.
\newblock \showarticletitle{Binarized neural networks}. In
  \bibinfo{booktitle}{\emph{NIPS}}.
\newblock


\bibitem[\protect\citeauthoryear{Iman, Helton, and Campbell}{Iman
  et~al\mbox{.}}{1981}]%
        {doi:10.1080/00224065.1981.11978748}
\bibfield{author}{\bibinfo{person}{Ronald~L. Iman}, \bibinfo{person}{Jon~C.
  Helton}, {and} \bibinfo{person}{James~E. Campbell}.}
  \bibinfo{year}{1981}\natexlab{}.
\newblock \showarticletitle{An Approach to Sensitivity Analysis of Computer
  Models: {Part} {I}—{Introduction}, Input Variable Selection and Preliminary
  Variable Assessment}.
\newblock \bibinfo{journal}{\emph{JQT}} \bibinfo{volume}{13},
  \bibinfo{number}{3} (\bibinfo{year}{1981}).
\newblock


\bibitem[\protect\citeauthoryear{Jackson}{Jackson}{2002}]%
        {Jackson01AlloyAlpha}
\bibfield{author}{\bibinfo{person}{Daniel Jackson}.}
  \bibinfo{year}{2002}\natexlab{}.
\newblock \showarticletitle{Alloy: A Lightweight Object Modeling Notation}.
\newblock \bibinfo{journal}{\emph{TOSEM}} \bibinfo{volume}{11},
  \bibinfo{number}{2} (\bibinfo{date}{April} \bibinfo{year}{2002}).
\newblock


\bibitem[\protect\citeauthoryear{Jackson and Sullivan}{Jackson and
  Sullivan}{2000}]%
        {JacksonS00}
\bibfield{author}{\bibinfo{person}{Daniel Jackson} {and}
  \bibinfo{person}{Kevin~J. Sullivan}.} \bibinfo{year}{2000}\natexlab{}.
\newblock \showarticletitle{{COM} revisited: Tool-assisted modelling of an
  architectural framework}. In \bibinfo{booktitle}{\emph{{SIGSOFT} {FSE}}}.
\newblock


\bibitem[\protect\citeauthoryear{Jackson and Vaziri}{Jackson and
  Vaziri}{2000}]%
        {JacksonVaziri00Bugs}
\bibfield{author}{\bibinfo{person}{Daniel Jackson} {and}
  \bibinfo{person}{Mandana Vaziri}.} \bibinfo{year}{2000}\natexlab{}.
\newblock \showarticletitle{Finding Bugs with a Constraint Solver}. In
  \bibinfo{booktitle}{\emph{ISSTA}}. \bibinfo{address}{Portland}.
\newblock


\bibitem[\protect\citeauthoryear{Johnstone}{Johnstone}{1979}]%
        {oldpaper}
\bibfield{author}{\bibinfo{person}{P.~T. Johnstone}.}
  \bibinfo{year}{1979}\natexlab{}.
\newblock \bibinfo{booktitle}{\emph{Conditions related to de Morgan's law}}.
\newblock


\bibitem[\protect\citeauthoryear{Katz, Barrett, Dill, Julian, and
  Kochenderfer}{Katz et~al\mbox{.}}{2017}]%
        {KaBaDiJuKo17Reluplex}
\bibfield{author}{\bibinfo{person}{G. Katz}, \bibinfo{person}{C. Barrett},
  \bibinfo{person}{D. Dill}, \bibinfo{person}{K. Julian}, {and}
  \bibinfo{person}{M. Kochenderfer}.} \bibinfo{year}{2017}\natexlab{}.
\newblock \showarticletitle{Reluplex: An Efficient {SMT} Solver for Verifying
  Deep Neural Networks}. In \bibinfo{booktitle}{\emph{CAV}}.
\newblock


\bibitem[\protect\citeauthoryear{Ke, Stolee, Goues, and Brun}{Ke
  et~al\mbox{.}}{2015}]%
        {Ke:2015:RPS:2916135.2916260}
\bibfield{author}{\bibinfo{person}{Yalin Ke}, \bibinfo{person}{Kathryn~T.
  Stolee}, \bibinfo{person}{Claire~Le Goues}, {and} \bibinfo{person}{Yuriy
  Brun}.} \bibinfo{year}{2015}\natexlab{}.
\newblock \showarticletitle{Repairing Programs with Semantic Code Search
  ({T})}. In \bibinfo{booktitle}{\emph{ASE}}.
\newblock


\bibitem[\protect\citeauthoryear{Khurshid and Jackson}{Khurshid and
  Jackson}{2000}]%
        {KhurshidJackson00ExploringDesignIntentional}
\bibfield{author}{\bibinfo{person}{Sarfraz Khurshid} {and}
  \bibinfo{person}{Daniel Jackson}.} \bibinfo{year}{2000}\natexlab{}.
\newblock \showarticletitle{Exploring the Design of an Intentional Naming
  Scheme with an Automatic Constraint Analyzer}. In
  \bibinfo{booktitle}{\emph{{ASE00}}}. \bibinfo{address}{Grenoble, France}.
\newblock


\bibitem[\protect\citeauthoryear{Kim and Kim}{Kim and Kim}{2011}]%
        {10.1007/978-3-642-24372-1_5}
\bibfield{author}{\bibinfo{person}{Moonzoo Kim} {and} \bibinfo{person}{Yunho
  Kim}.} \bibinfo{year}{2011}\natexlab{}.
\newblock \showarticletitle{Automated Analysis of Industrial Embedded
  Software}. In \bibinfo{booktitle}{\emph{Automated Technology for Verification
  and Analysis}}, \bibfield{editor}{\bibinfo{person}{Tevfik Bultan} {and}
  \bibinfo{person}{Pao-Ann Hsiung}} (Eds.). \bibinfo{publisher}{Springer Berlin
  Heidelberg}, \bibinfo{address}{Berlin, Heidelberg}, \bibinfo{pages}{51--59}.
\newblock
\showISBNx{978-3-642-24372-1}


\bibitem[\protect\citeauthoryear{{Korel}}{{Korel}}{1990}]%
        {57624}
\bibfield{author}{\bibinfo{person}{B. {Korel}}.}
  \bibinfo{year}{1990}\natexlab{}.
\newblock \showarticletitle{Automated software test data generation}.
\newblock \bibinfo{journal}{\emph{TSE}} \bibinfo{volume}{16},
  \bibinfo{number}{8} (\bibinfo{year}{1990}).
\newblock


\bibitem[\protect\citeauthoryear{Lagniez and Marquis}{Lagniez and
  Marquis}{2019}]%
        {ProjMC}
\bibfield{author}{\bibinfo{person}{Jean-Marie Lagniez} {and}
  \bibinfo{person}{Pierre Marquis}.} \bibinfo{year}{2019}\natexlab{}.
\newblock \showarticletitle{A Recursive Algorithm for Projected Model
  Counting}.
\newblock \bibinfo{journal}{\emph{AAAI}}  \bibinfo{volume}{33}
  (\bibinfo{year}{2019}), \bibinfo{pages}{1536--1543}.
\newblock


\bibitem[\protect\citeauthoryear{Liang, Bekker, and den Broeck}{Liang
  et~al\mbox{.}}{2017}]%
        {LiangETAL2017UAI}
\bibfield{author}{\bibinfo{person}{Yitao Liang}, \bibinfo{person}{Jessa
  Bekker}, {and} \bibinfo{person}{Guy~Van den Broeck}.}
  \bibinfo{year}{2017}\natexlab{}.
\newblock \showarticletitle{Learning the Structure of Probabilistic Sentential
  Decision Diagrams}. In \bibinfo{booktitle}{\emph{UAI}}.
\newblock


\bibitem[\protect\citeauthoryear{Long and Rinard}{Long and Rinard}{2016}]%
        {LongRinardPOPL2016}
\bibfield{author}{\bibinfo{person}{Fan Long} {and} \bibinfo{person}{Martin
  Rinard}.} \bibinfo{year}{2016}\natexlab{}.
\newblock \showarticletitle{Automatic Patch Generation by Learning Correct
  Code}. In \bibinfo{booktitle}{\emph{43rd Annual ACM SIGPLAN-SIGACT Symposium
  on Principles of Programming Languages (POPL)}}. \bibinfo{pages}{298--312}.
\newblock


\bibitem[\protect\citeauthoryear{Marinov and Khurshid}{Marinov and
  Khurshid}{2001}]%
        {MarinovKhurshid01TestEra}
\bibfield{author}{\bibinfo{person}{Darko Marinov} {and}
  \bibinfo{person}{Sarfraz Khurshid}.} \bibinfo{year}{2001}\natexlab{}.
\newblock \showarticletitle{{TestEra}: A novel framework for automated testing
  of Java programs}. In \bibinfo{booktitle}{\emph{ASE}}.
  \bibinfo{pages}{22--31}.
\newblock


\bibitem[\protect\citeauthoryear{Molina, Degiovanni, Ponzio, Regis, Aguirre,
  and Frias}{Molina et~al\mbox{.}}{2019}]%
        {MolinaETAL2019ICSE}
\bibfield{author}{\bibinfo{person}{Facundo Molina}, \bibinfo{person}{Renzo
  Degiovanni}, \bibinfo{person}{Pablo Ponzio}, \bibinfo{person}{German Regis},
  \bibinfo{person}{Nazareno Aguirre}, {and} \bibinfo{person}{Marcelo Frias}.}
  \bibinfo{year}{2019}\natexlab{}.
\newblock \showarticletitle{Training Binary Classifiers as Data Structure
  Invariants}. In \bibinfo{booktitle}{\emph{ICSE}}.
\newblock


\bibitem[\protect\citeauthoryear{Muhammad~Usman and Khurshid}{Muhammad~Usman
  and Khurshid}{2020}]%
        {TestMC2019}
\bibfield{author}{\bibinfo{person}{Wenxi~Wang Muhammad~Usman} {and}
  \bibinfo{person}{Sarfraz Khurshid}.} \bibinfo{year}{2020}\natexlab{}.
\newblock \bibinfo{title}{{TestMC}: A Framework for Testing Model Counters
  using Differential and Metamorphic Testing}.  (\bibinfo{year}{2020}).
\newblock
\newblock
\shownote{Under submission.}


\bibitem[\protect\citeauthoryear{Narodytska, Kasiviswanathan, Ryzhyk, Sagiv,
  and Walsh}{Narodytska et~al\mbox{.}}{2018}]%
        {NarodytskaKRSW18}
\bibfield{author}{\bibinfo{person}{Nina Narodytska},
  \bibinfo{person}{Shiva~Prasad Kasiviswanathan}, \bibinfo{person}{Leonid
  Ryzhyk}, \bibinfo{person}{Mooly Sagiv}, {and} \bibinfo{person}{Toby Walsh}.}
  \bibinfo{year}{2018}\natexlab{}.
\newblock \showarticletitle{Verifying Properties of Binarized Deep Neural
  Networks}. In \bibinfo{booktitle}{\emph{AAAI}}.
\newblock


\bibitem[\protect\citeauthoryear{Pei, Cao, Yang, and Jana}{Pei
  et~al\mbox{.}}{2017}]%
        {PeiETAL17}
\bibfield{author}{\bibinfo{person}{Kexin Pei}, \bibinfo{person}{Yinzhi Cao},
  \bibinfo{person}{Junfeng Yang}, {and} \bibinfo{person}{Suman Jana}.}
  \bibinfo{year}{2017}\natexlab{}.
\newblock \showarticletitle{{DeepXplore}: Automated Whitebox Testing of Deep
  Learning Systems}. In \bibinfo{booktitle}{\emph{{SOSP}}}.
\newblock


\bibitem[\protect\citeauthoryear{Quinlan}{Quinlan}{1987}]%
        {pathconditions}
\bibfield{author}{\bibinfo{person}{J.~R. Quinlan}.}
  \bibinfo{year}{1987}\natexlab{}.
\newblock \showarticletitle{Generating Production Rules from Decision Trees}.
  In \bibinfo{booktitle}{\emph{IJCAI}}.
\newblock


\bibitem[\protect\citeauthoryear{Rumbaugh, Jacobson, and Booch}{Rumbaugh
  et~al\mbox{.}}{1998}]%
        {RumbaughETAL98UML}
\bibfield{author}{\bibinfo{person}{J. Rumbaugh}, \bibinfo{person}{I. Jacobson},
  {and} \bibinfo{person}{G. Booch}.} \bibinfo{year}{1998}\natexlab{}.
\newblock \bibinfo{booktitle}{\emph{The Unified Modeling Language Reference
  Manual}}.
\newblock \bibinfo{publisher}{Addison-Wesley Object Technology Series}.
\newblock


\bibitem[\protect\citeauthoryear{Sahni and Gonzalez}{Sahni and
  Gonzalez}{1976}]%
        {Sahni:1976:PAP:321958.321975}
\bibfield{author}{\bibinfo{person}{Sartaj Sahni} {and} \bibinfo{person}{Teofilo
  Gonzalez}.} \bibinfo{year}{1976}\natexlab{}.
\newblock \showarticletitle{P-Complete Approximation Problems}.
\newblock \bibinfo{journal}{\emph{JACM}} \bibinfo{volume}{23},
  \bibinfo{number}{3} (\bibinfo{date}{July} \bibinfo{year}{1976}).
\newblock
\showISSN{0004-5411}
\urldef\tempurl%
\url{https://doi.org/10.1145/321958.321975}
\showDOI{\tempurl}


\bibitem[\protect\citeauthoryear{Samimi, Aung, and Millstein}{Samimi
  et~al\mbox{.}}{2010}]%
        {SamimiETALECOOP2010}
\bibfield{author}{\bibinfo{person}{Hesam Samimi}, \bibinfo{person}{Ei~Darli
  Aung}, {and} \bibinfo{person}{Todd~D. Millstein}.}
  \bibinfo{year}{2010}\natexlab{}.
\newblock \showarticletitle{Falling Back on Executable Specifications}. In
  \bibinfo{booktitle}{\emph{{ECOOP}}}.
\newblock


\bibitem[\protect\citeauthoryear{Sangal, Jordan, Sinha, and Jackson}{Sangal
  et~al\mbox{.}}{2005}]%
        {Sangal:2005:UDM:1103845.1094824}
\bibfield{author}{\bibinfo{person}{Neeraj Sangal}, \bibinfo{person}{Ev Jordan},
  \bibinfo{person}{Vineet Sinha}, {and} \bibinfo{person}{Daniel Jackson}.}
  \bibinfo{year}{2005}\natexlab{}.
\newblock \showarticletitle{Using Dependency Models to Manage Complex Software
  Architecture}.
\newblock \bibinfo{journal}{\emph{SIGPLAN Not.}} \bibinfo{volume}{40},
  \bibinfo{number}{10} (\bibinfo{date}{Oct.} \bibinfo{year}{2005}),
  \bibinfo{pages}{167--176}.
\newblock
\showISSN{0362-1340}
\urldef\tempurl%
\url{https://doi.org/10.1145/1103845.1094824}
\showDOI{\tempurl}


\bibitem[\protect\citeauthoryear{Shalev-Shwartz, Shamir, Srebro, and
  Sridharan}{Shalev-Shwartz et~al\mbox{.}}{2010}]%
        {Shalev-Shwartz:2010:LSU:1756006.1953019}
\bibfield{author}{\bibinfo{person}{Shai Shalev-Shwartz}, \bibinfo{person}{Ohad
  Shamir}, \bibinfo{person}{Nathan Srebro}, {and} \bibinfo{person}{Karthik
  Sridharan}.} \bibinfo{year}{2010}\natexlab{}.
\newblock \showarticletitle{Learnability, Stability and Uniform Convergence}.
\newblock \bibinfo{journal}{\emph{JMLR}}  \bibinfo{volume}{11}
  (\bibinfo{date}{Dec.} \bibinfo{year}{2010}).
\newblock
\showISSN{1532-4435}
\urldef\tempurl%
\url{http://dl.acm.org/citation.cfm?id=1756006.1953019}
\showURL{%
\tempurl}


\bibitem[\protect\citeauthoryear{Shih, Choi, and Darwiche}{Shih
  et~al\mbox{.}}{2018}]%
        {ShihETAL2018}
\bibfield{author}{\bibinfo{person}{Andy Shih}, \bibinfo{person}{Arthur Choi},
  {and} \bibinfo{person}{Adnan Darwiche}.} \bibinfo{year}{2018}\natexlab{}.
\newblock \showarticletitle{Formal Verification of Bayesian Network
  Classifiers}. In \bibinfo{booktitle}{\emph{PGM}}.
\newblock


\bibitem[\protect\citeauthoryear{Shlyakhter}{Shlyakhter}{2001}]%
        {Shlyakhter01EffectiveSymmetryBreaking}
\bibfield{author}{\bibinfo{person}{Ilya Shlyakhter}.}
  \bibinfo{year}{2001}\natexlab{}.
\newblock \showarticletitle{Generating Effective Symmetry-Breaking Predicates
  for Search Problems}. In \bibinfo{booktitle}{\emph{SAT}}.
\newblock


\bibitem[\protect\citeauthoryear{Si, Dai, Raghothaman, Naik, and Song}{Si
  et~al\mbox{.}}{2018}]%
        {Si:2018:LLI:3327757.3327873}
\bibfield{author}{\bibinfo{person}{Xujie Si}, \bibinfo{person}{Hanjun Dai},
  \bibinfo{person}{Mukund Raghothaman}, \bibinfo{person}{Mayur Naik}, {and}
  \bibinfo{person}{Le Song}.} \bibinfo{year}{2018}\natexlab{}.
\newblock \showarticletitle{Learning Loop Invariants for Program Verification}.
  In \bibinfo{booktitle}{\emph{NIPS}}.
\newblock


\bibitem[\protect\citeauthoryear{Solar-Lezama, Rabbah, Bod{\'\i}k, and
  Ebcio{\u{g}}lu}{Solar-Lezama et~al\mbox{.}}{2005}]%
        {Sketching}
\bibfield{author}{\bibinfo{person}{Armando Solar-Lezama},
  \bibinfo{person}{Rodric Rabbah}, \bibinfo{person}{Rastislav Bod{\'\i}k},
  {and} \bibinfo{person}{Kemal Ebcio{\u{g}}lu}.}
  \bibinfo{year}{2005}\natexlab{}.
\newblock \showarticletitle{Programming by sketching for bit-streaming
  programs}. In \bibinfo{booktitle}{\emph{ACM SIGPLAN Conference on Programming
  Language Design and Implementation (PLDI)}}. \bibinfo{pages}{281--294}.
\newblock


\bibitem[\protect\citeauthoryear{Soos and Meel}{Soos and Meel}{2019}]%
        {SM19}
\bibfield{author}{\bibinfo{person}{Mate Soos} {and} \bibinfo{person}{Kuldeep~S.
  Meel}.} \bibinfo{year}{2019}\natexlab{}.
\newblock \showarticletitle{{BIRD}: Engineering an Efficient {CNF}-{XOR} {SAT}
  Solver and its Applications to Approximate Model Counting}. In
  \bibinfo{booktitle}{\emph{AAAI}}.
\newblock


\bibitem[\protect\citeauthoryear{Spivey}{Spivey}{1992}]%
        {Spivey92Z}
\bibfield{author}{\bibinfo{person}{J.~M. Spivey}.}
  \bibinfo{year}{1992}\natexlab{}.
\newblock \bibinfo{booktitle}{\emph{The {Z} Notation: A Reference Manual}
  (\bibinfo{edition}{second} ed.)}.
\newblock \bibinfo{publisher}{Prentice Hall}.
\newblock


\bibitem[\protect\citeauthoryear{Sun, Wu, Ruan, Huang, Kwiatkowska, and
  Kroening}{Sun et~al\mbox{.}}{2018}]%
        {SunETAL2018ASE}
\bibfield{author}{\bibinfo{person}{Youcheng Sun}, \bibinfo{person}{Min Wu},
  \bibinfo{person}{Wenjie Ruan}, \bibinfo{person}{Xiaowei Huang},
  \bibinfo{person}{Marta Kwiatkowska}, {and} \bibinfo{person}{Daniel
  Kroening}.} \bibinfo{year}{2018}\natexlab{}.
\newblock \showarticletitle{Concolic Testing for Deep Neural Networks}. In
  \bibinfo{booktitle}{\emph{ASE}}.
\newblock


\bibitem[\protect\citeauthoryear{Tian, Pei, Jana, and Ray}{Tian
  et~al\mbox{.}}{2018}]%
        {TianETAL18DeepTest}
\bibfield{author}{\bibinfo{person}{Yuchi Tian}, \bibinfo{person}{Kexin Pei},
  \bibinfo{person}{Suman Jana}, {and} \bibinfo{person}{Baishakhi Ray}.}
  \bibinfo{year}{2018}\natexlab{}.
\newblock \showarticletitle{{DeepTest}: Automated testing of
  deep-neural-network-driven autonomous cars}. In
  \bibinfo{booktitle}{\emph{ICSE}}.
\newblock


\bibitem[\protect\citeauthoryear{Trippel, Lustig, and Martonosi}{Trippel
  et~al\mbox{.}}{2018}]%
        {CheckMateMICRO2018}
\bibfield{author}{\bibinfo{person}{Caroline Trippel}, \bibinfo{person}{Daniel
  Lustig}, {and} \bibinfo{person}{Margaret Martonosi}.}
  \bibinfo{year}{2018}\natexlab{}.
\newblock \showarticletitle{{CheckMate}: Automated Synthesis of Hardware
  Exploits and Security Litmus Tests}. In \bibinfo{booktitle}{\emph{MICRO}}.
\newblock


\bibitem[\protect\citeauthoryear{Trippel, Lustig, and Martonosi}{Trippel
  et~al\mbox{.}}{2019}]%
        {CheckMateMicro2019}
\bibfield{author}{\bibinfo{person}{Caroline Trippel}, \bibinfo{person}{Daniel
  Lustig}, {and} \bibinfo{person}{Margaret Martonosi}.}
  \bibinfo{year}{2019}\natexlab{}.
\newblock \showarticletitle{Security Verification via Automatic Hardware-Aware
  Exploit Synthesis: The {CheckMate} Approach}.
\newblock \bibinfo{journal}{\emph{{IEEE} Micro}} \bibinfo{volume}{39},
  \bibinfo{number}{3} (\bibinfo{year}{2019}).
\newblock


\bibitem[\protect\citeauthoryear{Tseytin}{Tseytin}{1966}]%
        {Tseytin1966}
\bibfield{author}{\bibinfo{person}{G.~S. Tseytin}.}
  \bibinfo{year}{1966}\natexlab{}.
\newblock \bibinfo{title}{On the complexity of derivation in propositional
  calculus.}  (\bibinfo{year}{1966}).
\newblock
\newblock
\shownote{Presented at the Leningrad Seminar on Mathematical Logic.}


\bibitem[\protect\citeauthoryear{Usman, Wang, Wang, Yelen, Dini, and
  Khurshid}{Usman et~al\mbox{.}}{2019}]%
        {UsmanETAL2019SPIN}
\bibfield{author}{\bibinfo{person}{Muhammad Usman}, \bibinfo{person}{Wenxi
  Wang}, \bibinfo{person}{Kaiyuan Wang}, \bibinfo{person}{Cagdas Yelen},
  \bibinfo{person}{Nima Dini}, {and} \bibinfo{person}{Sarfraz Khurshid}.}
  \bibinfo{year}{2019}\natexlab{}.
\newblock \showarticletitle{A Study of Learning Data Structure Invariants Using
  Off-the-shelf Tools}. In \bibinfo{booktitle}{\emph{SPIN}}.
\newblock


\bibitem[\protect\citeauthoryear{Valiant}{Valiant}{1984}]%
        {Valiant:1984:TL:1968.1972}
\bibfield{author}{\bibinfo{person}{L.~G. Valiant}.}
  \bibinfo{year}{1984}\natexlab{}.
\newblock \showarticletitle{A Theory of the Learnable}.
\newblock \bibinfo{journal}{\emph{CACM}} \bibinfo{volume}{27},
  \bibinfo{number}{11} (\bibinfo{date}{Nov.} \bibinfo{year}{1984}).
\newblock
\showISSN{0001-0782}
\urldef\tempurl%
\url{https://doi.org/10.1145/1968.1972}
\showDOI{\tempurl}


\bibitem[\protect\citeauthoryear{Vapnik}{Vapnik}{1995}]%
        {Vapnik:1995:NSL:211359}
\bibfield{author}{\bibinfo{person}{Vladimir~N. Vapnik}.}
  \bibinfo{year}{1995}\natexlab{}.
\newblock \bibinfo{booktitle}{\emph{The Nature of Statistical Learning
  Theory}}.
\newblock
\showISBNx{0-387-94559-8}


\bibitem[\protect\citeauthoryear{Vapnik and Ya.~Chervonenkis}{Vapnik and
  Ya.~Chervonenkis}{1971}]%
        {article1971a}
\bibfield{author}{\bibinfo{person}{V.~N. Vapnik} {and} \bibinfo{person}{A
  Ya.~Chervonenkis}.} \bibinfo{year}{1971}\natexlab{}.
\newblock \showarticletitle{On the Uniform Convergence of Relative Frequencies
  of Events to Their Probabilities}.
\newblock \bibinfo{journal}{\emph{Theory of Probabibility and its
  Applicactions}} (\bibinfo{year}{1971}).
\newblock


\bibitem[\protect\citeauthoryear{Vasic, Petrovic, Wang, Nikolic, Singh, and
  Khurshid}{Vasic et~al\mbox{.}}{2019}]%
        {VasicETAL19MOET}
\bibfield{author}{\bibinfo{person}{Marko Vasic}, \bibinfo{person}{Andrija
  Petrovic}, \bibinfo{person}{Kaiyuan Wang}, \bibinfo{person}{Mladen Nikolic},
  \bibinfo{person}{Rishabh Singh}, {and} \bibinfo{person}{Sarfraz Khurshid}.}
  \bibinfo{year}{2019}\natexlab{}.
\newblock \showarticletitle{Mo{\"{E}}T: Interpretable and Verifiable
  Reinforcement Learning via Mixture of Expert Trees}.
\newblock \bibinfo{journal}{\emph{CoRR}} (\bibinfo{year}{2019}).
\newblock


\bibitem[\protect\citeauthoryear{Wang, Usman, Almaawi, Wang, Meel, and
  Khurshid}{Wang et~al\mbox{.}}{2020}]%
        {StudySymmetry}
\bibfield{author}{\bibinfo{person}{Wenxi Wang}, \bibinfo{person}{Muhammad
  Usman}, \bibinfo{person}{Alyas Almaawi}, \bibinfo{person}{Kaiyuan Wang},
  \bibinfo{person}{Kuldeep~S. Meel}, {and} \bibinfo{person}{Sarfraz Khurshid}.}
  \bibinfo{year}{2020}\natexlab{}.
\newblock \showarticletitle{A Study of Symmetry Breaking Predicates and Model
  Counting}. In \bibinfo{booktitle}{\emph{TACAS}}.
\newblock
\newblock
\shownote{To appear.}


\bibitem[\protect\citeauthoryear{Wickerson, Batty, Sorensen, and
  Constantinides}{Wickerson et~al\mbox{.}}{2017}]%
        {WickersonETAL2017}
\bibfield{author}{\bibinfo{person}{John Wickerson}, \bibinfo{person}{Mark
  Batty}, \bibinfo{person}{Tyler Sorensen}, {and} \bibinfo{person}{George~A.
  Constantinides}.} \bibinfo{year}{2017}\natexlab{}.
\newblock \showarticletitle{Automatically Comparing Memory Consistency Models}.
  In \bibinfo{booktitle}{\emph{POPL}}.
\newblock


\bibitem[\protect\citeauthoryear{Zaeem and Khurshid}{Zaeem and
  Khurshid}{2010}]%
        {ZaeemKhurshidECOOP2010}
\bibfield{author}{\bibinfo{person}{Razieh~Nokhbeh Zaeem} {and}
  \bibinfo{person}{Sarfraz Khurshid}.} \bibinfo{year}{2010}\natexlab{}.
\newblock \showarticletitle{Contract-Based Data Structure Repair Using
  {Alloy}}. In \bibinfo{booktitle}{\emph{{ECOOP}}}.
\newblock


\bibitem[\protect\citeauthoryear{Zave}{Zave}{2012}]%
        {ZaveChrod2012}
\bibfield{author}{\bibinfo{person}{Pamela Zave}.}
  \bibinfo{year}{2012}\natexlab{}.
\newblock \showarticletitle{Using Lightweight Modeling to Understand {Chord}}.
\newblock \bibinfo{journal}{\emph{SIGCOMM CCR}} \bibinfo{volume}{42},
  \bibinfo{number}{2} (\bibinfo{year}{2012}).
\newblock


\bibitem[\protect\citeauthoryear{{Zave}}{{Zave}}{2017}]%
        {ZaveChordTSE}
\bibfield{author}{\bibinfo{person}{P. {Zave}}.}
  \bibinfo{year}{2017}\natexlab{}.
\newblock \showarticletitle{Reasoning About Identifier Spaces: How to Make
  Chord Correct}.
\newblock \bibinfo{journal}{\emph{IEEE Transactions on Software Engineering}}
  (\bibinfo{year}{2017}).
\newblock


\end{thebibliography}
\end{document}